\documentclass[letterpaper, paper,10pt]{AAS}		

\usepackage{bm}
\usepackage{amsmath}
\usepackage{subfigure}
\usepackage[colorlinks=true, pdfstartview=FitV, linkcolor=black, citecolor= black, urlcolor= black]{hyperref}
\usepackage{overcite}
\usepackage{footnpag}			      	
\graphicspath{{./figures/}}
\usepackage{amssymb}
\newcommand{\argmaxF}{\mathop{\mathrm{argmax}}\limits}   
\usepackage{caption}
\PaperNumber{21-421}
\DeclareMathOperator*{\argmin}{arg\,min}
\usepackage{booktabs}

\begin{document}


\title{Safe and Uncertainty-Aware Robotic Motion Planning Techniques for Agile On-Orbit Assembly}

\author{Bryce Doerr\thanks{Postdoctoral Fellow, Department of Aeronautics and Astronautics, Massachusetts Institute of Technology, 02139.},
Keenan Albee\thanks{Ph.D. Student, Department of Aeronautics and Astronautics, Massachusetts Institute of Technology, 02139},
Monica Ekal\thanks{Ph.D. Student, Institute for Systems and Robotics, Instituto Superior Técnico, 1049-001}, 
Richard Linares\thanks{Charles Stark Draper Assistant Professor, Department of Aeronautics and Astronautics, Massachusetts Institute of Technology, 02139.}, \\ and Rodrigo Ventura\thanks{Assistant Professor, Institute for Systems and Robotics, Instituto Superior Técnico, 1049-001}
}

\maketitle{}

\begin{abstract}
As access to space and robotic autonomy capabilities move forward, there is simultaneously a growing interest in deploying large, complex space structures to provide new on-orbit capabilities. New space-borne observatories, large orbital outposts, and even futuristic on-orbit manufacturing will be enabled by robotic assembly of space structures using techniques like on-orbit additive manufacturing which can provide flexibility in constructing and even repairing complex hardware. However, the dynamics underlying the robotic assembly system during manipulation may operate under uncertainties (e.g. changing inertial properties). Thus, inertial estimation of the robotic assembler and the manipulated additively manufactured component must be considered during the structural assembly process. The contribution of this work is to address both the motion planning and control for robotic assembly with consideration of the inertial estimation of the combined free-flying robotic assembler and additively manufactured component system. Specifically, the Linear Quadratic Regulator Rapidly-Exploring Randomized Trees (LQR-RRT*) and dynamically feasible path smoothing are used to obtain obstacle-free trajectories for the system. Further, model learning is incorporated explicitly into the planning stages via approximation of the continuous system and accompanying reward of performing safe, objective-oriented motion. Remaining uncertainty can then be dealt with explicitly via robust tube model predictive control techniques. By obtaining controlled trajectories that consider both obstacle avoidance and learning of the inertial properties of the free-flyer and manipulated component system, the free-flyer rapidly considers and plans the construction of space structures with enhanced system knowledge. The approach naturally generalizes to repairing, refueling, and re-provisioning space structure components while providing optimal collision-free trajectories under e.g., inertial uncertainty.
\end{abstract}

\section{Introduction}
On-orbit robotic assembly of large, complex space structures is an emerging area of autonomy that can provide greater access to scientific, communicative, and observational knowledge that is otherwise limited or unknown. Currently, there are many Earth observing satellites investigating the Earth’s oceans and land masses (e.g. Landsat-7 \cite{goward2001landsat}), and there is a need to increase the longevity of these missions by on-orbit servicing and assembly. Specifically, new servicing and assembly technologies are being developed through NASA's On-orbit Servicing, Assembly, and Manufacturing 1 (OSAM-1) servicing mission which will refuel Landsat 7 in 2022 \cite{reed2016restore,coll2020satellite}. On-orbit servicing is a superset of areas including repairing, refueling, and re-provisioning of spacecraft \cite{saleh2003flexibility}. Concepts in the area of robotic assembly of space structures have also been proposed by industry and academia. Tethers Unlimited, Inc. is enabling the area of on-orbit fabrication including antennas, solar panels, and truss structures using SpiderFab \cite{hoyt2013spiderfab}. Made In Space, Inc. is developing the manufacturing and assembly of spacecraft components on-orbit using its Archinaut One concept \cite{patane2017archinaut}. These assemblers are defined by capturing the structure with a robotic arm and climbing the structure for servicing or construction. Alternatively, space structures can be assembled using proximity operations with free-flyer robots, providing benefits in terms of mission flexibility and range of motion \cite{jewison2014definition}. At MIT's Space Systems Laboratory (SSL), Astrobee, a six degree of freedom (DOF) free-flyer with a 3 DOF robotic arm, is being used to develop capabilities relating to autonomy, multi-agent coordination, and microgravity manipulation. This robotic system is also available as a testbed on the International Space Station (ISS) \cite{bualat2015astrobee}. Thus, Astrobee has the functionality necessary to serve as a testbed for developing modular motion planning strategies for on-orbit assembly. By enabling the autonomous robotic assembly of space structures, near-Earth science can be improved (e.g. upgrading new components) and its life can be extended (e.g. repairing, refueling, and re-provisioning). 

To further make on-orbit assembly a reality, significant progress has been made in additive manufacturing. For example, geometrically complex components and devices can be produced using improvements in additive manufacturing \cite{rosen2007computer,schaedler2016architected}. Currently in the on-orbit environment, a 3D printer is in operation on the ISS since 2014 \cite{thomas2017effect}. This printer facility allows for production of 3D thermopolymers for the use of components and tools for repairs and upgrades of existing hardware. These additive manufacturing architectures can be naturally extended to manufacturing components for a space structure which can be assembled autonomously while also providing capabilities for repairing and re-provisioning of the space structure. Thus, this technology has the potential to effectively maintain and upgrade the space structure.

In the area of robotic assembly, modular control strategies utilizing component geometry have been developed to build structures using ground and aerial robotics \cite{petersen2011termes,willmann2012aerial}. Collaborative multi-robotic systems have also been designed to transport and manipulate voxels to construct plates, enclosures, and cellular beams, \cite{jenett2019material} and through coarse and fine manipulation techniques \cite{dogar2015multi}. Distributed control has been used for larger multi-robotic systems so agents can collectively climb and assemble the structure they are building on \cite{werfel2014designing}. Significant progress has also been made for on-orbit assembly strategies. Superquadric potential fields have been applied to assemble components of a structure while providing collision avoidance \cite{badawy2008orbit}. Other methods involve using innovative techniques from the ground robotics field which include implementation of sampling-based motion planning using Linear Quadratic Regulator Rapidly-Exploring Randomized Trees (LQR-RRT*), trajectory smoothing via shortcutting, and Nonlinear Model Predictive Control (NMPC) \cite{perez2012lqr,geraerts2007creating,sathya2018embedded,doerr2020motion}. Particularly with this method, rapid consideration and planning for obtaining collision-free trajectories are found in a state-space that increasingly becomes more complex as structural components are produced.


On-orbit motion planning for microgravity proximity operations is not immune to uncertainty, including uncertainties that arise due to system parameter knowledge, environmental awareness, and sensor accuracy. A number of techniques exist to either explicitly account for and resolve uncertainty, which can increase the knowledge of the system or provide guarantees on performance when faced with certain inherent, unresolvable uncertainty. Some forms of uncertainty are \textit{parametric} and can be resolved using knowledge of the system model. For example, if a free-flyer rigidly grasps a target, the underyling rigid body dynamics do not change, but certain \textit{parameters} in the system model are modified. On-orbit manufactured objects in particular may not be fully inertially characterized.

Existing planning and control under uncertainty approaches are wide-ranging, but can be broadly put into two categories with respect to system model information gain. Some approaches attempt complete system identification (sys ID) followed by trajectory planning with the estimated nominal model \cite{lampariello2005modeling} or otherwise are robust or chance-constrained approaches that operate under an assumed uncertainty bound \cite{How2001,majumdar2017funnel,lopez2019dynamic}. More exotic approaches attempt to consider the parameter learning problem directly, sometimes combined with accounting for control in the the planning. For example, some work involving POMDPs with covariance minimization in the cost function \cite{Webb2014}, and some very recent work on covariance steering \cite{Okamoto2018,Okamoto2019} exists. However, these approaches have not been implemented on hardware, and their scalability has not yet been demonstrated. Control methods are more concerned with performance of tracking or regulating existing trajectories but sometimes consider online model adaptation, including adaptive control \cite{Slotine,espinoza2017concurrent}, and self-tuning (ST) control \cite{xu1994parameterization}, whereas other approaches are more interested in robustness guarantees.

For a coupled on-orbit free-flyer and manufactured component, inertial parameter estimation is necessary to provide accurate information of the combined system for motion planning and control. Thus, optimal planning over parameters such as fuel efficiency and precise execution using model-based control can be obtained after adequately characterizing the dynamically coupled system. When inertial parameters change during operation, online methods of estimation can be used. This is especially relevant for the aforementioned on-orbit assembly operations, where robotic free-flyers would need to safely transport various components to their assembly locations, and precision and safety is paramount.

The most common class of rigid-body on-orbit identification methods for space robotic systems involve either the use of Newton-Euler (NE) dynamic equations \cite{keim2006spacecraft,murotsu1994parameter,ekal2018inertial} or the property of conservation of angular momentum \cite{yoshida2002inertia,ma2008orbit,christidi2017}. In both cases, the dynamic equations are rearranged such that the unknown parameters appear linearly. While the NE equations require acceleration measurements, which are noisy, obtaining a linear formulation with the angular-momentum method is at times not straightforward. 
In order to ensure that the regressor matrix is full-rank and thus invertible, the robot is made to track information rich, exciting trajectories. Batch or recursive least squares approaches can then be used to yield the estimates. Using the discrete variational equations of motion is a recent addition to this category of methods, as it does not need acceleration data unlike the NE method and could result in greater estimate accuracy as compared to finite difference discretizations of the NE equations \cite{manchester2017recursive,ekal2020dual}.

Sequential filtering methods such as the Kalman Filter or Unscented Kalman Filter have been used for either parameter estimation, or joint estimation of states and parameters \cite{lichter2004state,vandyke2004unscented}. Apart from potentially being real-time, the possibility of fusing information from multiple sensors is an added benefit in these methods.
Adaptive or concurrent learning has been proposed to perform system identification in conjunction with adaptive control under the condition of persistent excitation
\cite{espinoza2017concurrent,shin1994}.
Finally, the assumption of rigid bodies does not always hold when it comes to space applications. Estimating properties of spacecraft with flexible appendages and generating appropriate excitation maneuvers has been addressed \cite{rackl2014parameter,nanos2019parameter}. Estimation of flexible structures like trusses and beams, which could be pertinent to an on-orbit assembly problem, has been previously handled by using the finite element method \cite{liu2018modeling} or distributed parameter models \cite{taylor1991parameter}, among others.

This work addresses the motion planning and control for on-orbit robotic assembly in the presence of inertial parametric uncertainty due to object manipulation. By extending LQR-RRT*, trajectory smoothing via LQR shortcutting, obstacle avoidance, inertial model learning, and robust tube model predictive control (MPC), to the Astrobee testbed, safe and uncertainty-aware techniques can be obtained for on-orbit robotic assembly of space structures, which can be expanded to constructing next generation telescopes, space stations, and communication satellites.

\section{Problem Formulation}
\begin{figure}[h]
\begin{centering}
      \includegraphics[width=1\textwidth]{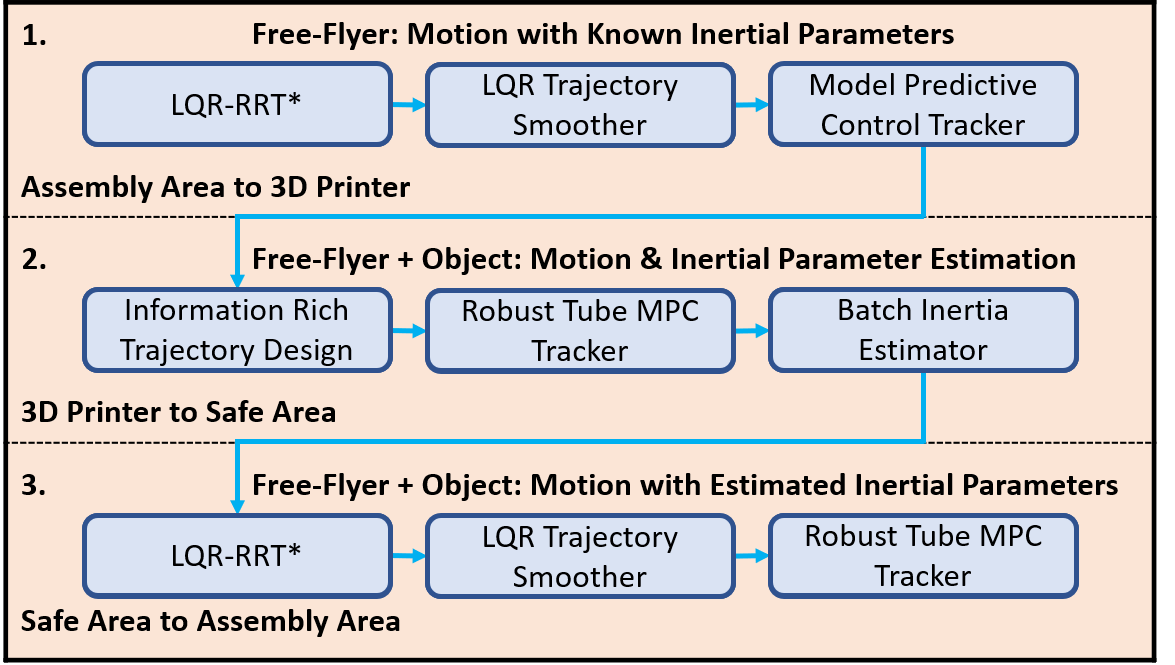}
        \caption{A block diagram of the safe and uncertainty-aware robotic motion planning architecture. }\label{planblockdiagram}
\end{centering}
\end{figure}

This work uses LQR-RRT*, trajectory smoothing via LQR shortcutting, inertial model learning using information gain, and robust tube MPC to provide safe and uncertainty-aware robotic motion planning for on-orbit assembly of space structures. Figure \ref{planblockdiagram} shows the architecture designed to plan and control a robotic free-flyer under uncertain inertial parameters due to object manipulation for on-orbit assembly. The planning architecture contains three specific segments as part of the assembly process. In the first segment, the free-flyer with known inertial parameters plans its trajectory from the assembly area to the 3D printer. LQR-RRT* is applied to the free-flyer to obtain an initial sub-optimal and collision-free trajectory, although it has capabilities on reaching asymptotically optimal trajectories as the state-space is sampled to infinity. The initial sub-optimal trajectory may be jerky and unnatural with respect to the target due to the sampling-based nature of the algorithm, so trajectory smoothing using LQR shortcutting is applied to reduce the effect of ``randomness'' from the LQR-RRT* trajectory while also maintaining collision-free trajectories. Lastly, MPC is used to track the trajectory of the free-flyer with known inertial parameters. Once the free-flyer reaches the 3D printer and manipulates the 3D printed object, the inertial parameters of the free-flyer and object system become uncertain. This leads to the second segment of identifying the inertial parameters of the free-flyer and object system. The goal is to obtain information rich (exciting) trajectories to characterize the inertial parameters for the system. To facilitate convergence of inertial estimates, an information-rich trajectory is obtained between the 3D printer and a safe area (no obstacles) by maximizing the trace of the Fisher information over the trajectory. Next, robust tube MPC is used to track this trajectory for an assumed bounded inertial uncertainty. The system response obtained through execution of this trajectory is used by a batch maximum likelihood estimator to obtain inertial estimates at the safe area. With the updated system model, LQR-RRT*, trajectory smoothing by LQR shortcutting, and robust tube MPC can be used to plan and control the free-flyer and object system from the safe area to the assembly area for construction. Robust tube MPC is used again to consider any remaining uncertainty in the system to provide safe motion during assembly. To begin, background for motion planning, control, and estimation is discussed for a nonlinear system.

A free-flyer robot with a continuous-time nonlinear dynamics is given by
\begin{equation}\label{fxu}
    \dot{\mathbf{x}}(t)=f(\mathbf{x}(t),\mathbf{u}(t),\boldsymbol{\theta}(t)),
\end{equation}
where $\mathbf{x}(t)\in\mathbb{R}^{n_{\mathbf{x}}}$ is the robot state, $\mathbf{u}(t)\in\mathbb{R}^{n_{\mathbf{u}}}$ is the robot control input, and $\boldsymbol{\theta}(t)\in\mathbb{R}^{n_{\boldsymbol{\theta}}}$ about time $t$ where $n_{\mathbf{x}}$, $n_{\mathbf{u}}$, and $n_{\boldsymbol{\theta}}$ are the sizes of the state, control, and inertia parameter vectors, respectively. The state vector contains the translational and rotational motion that describes the robot as well as the motion that describes any manipulator attached.  The control vector contains the forces and torques necessary to actuate the robot and any manipulator joints. The inertial parameters contain the mass and moments of inertia for the free-flyer robot.

Motion planning and control using LQR-RRT*, trajectory smoothing via LQR shortcutting, and robust tube MPC are defined for discrete dynamics and require linear time-varying (LTV) system approximations. By a first-order Taylor series expansion around an operating point for linearization \cite{markley2014fundamentals} and a discretization using a fourth-order Runge-Kutta method with a zero-order hold on the control input \cite{van1978computing}, an LTV system that approximates the nonlinear system is given by
\begin{equation}\label{axbu}
    \delta\mathbf{x}_{k+1}=A_k\delta\mathbf{x}_k+B_k\delta\mathbf{u}_k,
\end{equation}
where  $A_k$ and $B_k$ are the state and control matrices at a time-step $k$. The terms $\delta\mathbf{x}_k=\left( \mathbf{x}_k-\bar{\mathbf{x}}_k\right)$ and $\delta\mathbf{u}_k=\left( \mathbf{u}_k-\bar{\mathbf{u}}_k\right)$ are the perturbation state and control about some operating point $\bar{\mathbf{x}}_k$. For this work, the operating point occurs at some target $\bar{\mathbf{x}}_k=\mathbf{x}_{des}$.

The objective is to plan and control the robotic free-flyer about a quadratic cost function
\begin{equation}\label{costlqr}
J(\delta{\bf x}_0,\delta U_{0:N-1})=\sum_{k=0}^{N-1}l(\delta{\bf x}_k,\delta{\bf u}_k)+l_N(\delta{\bf x}_N),
\end{equation}
where $\delta U_{0:N-1}=\left[\delta{\bf u}_{0},\delta{\bf u}_{1},\cdots,\delta{\bf u}_{N-1}\right]$ is the control sequence, $l(\delta{\bf x}_k,\delta{\bf u}_k)$ is the running cost, and $l_N(\delta{\bf x}_N)$ is the terminal cost to a time $N$. This is given by
\begin{equation}\label{immediateterminallqr}
l(\delta{\bf x}_k,\delta{\bf u}_k)=\frac{1}{2}\left[ \begin{array}{c} 1 \\\delta { \bf x}_k \\ \delta\mathbf{u}_k \end{array} \right]^{T} \begin{bmatrix} 0& \mathbf{q}_{k}^{T} & \mathbf{r}_{k}^{T} \\ \mathbf{q}_{k}&Q_{k} & P_{k} \\\mathbf{r}_{k}&P_{k} & R_{k}  \end{bmatrix}\left[ \begin{array}{c} 1\\ \delta{ \bf x}_k \\ \delta\mathbf{u}_k \end{array} \right],\hspace{12pt}l_N(\delta{\bf x}_N)=\frac{1}{2}\delta{ \bf x}_N ^{T}Q_{N}\delta{ \bf x}_N+\delta{ \bf x}_N ^{T}\mathbf{q}_{N},
\end{equation}
where $\mathbf{q}_{k}$, $\mathbf{r}_{k}$, $Q_{k}$, $R_{k}$, and $P_{k}$ are the running weights (coefficients), and $Q_{N}$ and $\mathbf{q}_{N}$ are the terminal weights. The weight matrices, $Q_{k}$ and $R_{k}$, are positive definite and the block matrix $\begin{bmatrix} Q_{k} & P_{k} \\P_{k} & R_{k}  \end{bmatrix}$ is positive-semidefinite \cite{13}. This is constrained to dynamics and obstacles (structural components) of the assembly problem itself.

\section{Sampling-Based Motion Planning}
The motion planning for on-orbit assembly using a free-flyer is developed using LQR-RRT* and trajectory smoothing using LQR shortcutting. The motivation of using sampling-based motion planning is that it provides computationally efficient, collision-free trajectories through random sampling of the complex state-space \cite{perez2012lqr}. In literature, LQR-RRT* and trajectory smoothing have been discussed extensively \cite{paden2016survey,geraerts2007creating,kallmann2008planning}. In this work, LQR-RRT* and trajectory smoothing using LQR shortcutting are discussed with application to on-orbit assembly using free-flyers which follows from previous work \cite{perez2012lqr,doerr2020motion}.
\subsection{LQR-RRT*}
For a nonlinear system given in Eq. \eqref{fxu}, the goal is to obtain a trajectory that minimizes the quadratic cost function given in Eqs. \eqref{costlqr} and \eqref{immediateterminallqr} with initial and final states, $\mathbf{x}_0$ and $\mathbf{x}_{des}$, respectively. The quadratic cost is simplified to
\begin{equation}
\begin{gathered}
J(\delta{\bf x}_0,\delta U_{0:N-1})=\sum_{k=0}^{N-1}\delta{ \bf x}_k ^{T}\mathbf{q}_{k}+\delta{ \bf u}_k^{T}\mathbf{r}_{k}+\frac{1}{2}\delta{ \bf x}_k ^{T}Q_{k}\delta{ \bf x}_k+\frac{1}{2}\delta{ \bf u}_k ^{T}R_{k}\delta{ \bf u}_k +\delta{ \bf u}_k ^{T}P_{k}\delta{ \bf x}_k
\\+\frac{1}{2}\delta{ \bf x}_N ^{T}Q_{N}\delta{ \bf x}_N+\delta{ \bf x}_N ^{T}\mathbf{q}_{N}.
\end{gathered}
\end{equation}
The optimal solution with respect to the cost function in terms of the control sequence is given by
\begin{equation}\label{minJ}
\delta U_{0:N-1}^{\star}(\delta{\bf x}_0)=\argmin_{\delta U_{0:N-1}} J(\delta{\bf x}_0,\delta U_{0:N-1}).
\end{equation}
To solve for the control solution in Eq. \eqref{minJ}, a value iteration is used which determines the optimal cost-to-go (value) starting from the final time-step and moving backwards in time minimizing the control sequence. This is given by
\begin{subequations}
\begin{equation}\label{costgo}
J(\delta{\bf x}_{k},\delta U_{k:N-1})=\sum_{k}^{N-1}l(\delta{\bf x}_k,\delta{\bf u}_k)+l_N(\delta{\bf x}_N),
\end{equation}
\begin{equation}\label{valuefunction}
V(\delta{\bf x}_{k})=\min_{\delta U_{k:N-1}} J(\delta{\bf x}_{k},\delta U_{k:N-1}),
\end{equation}
\end{subequations}
which is similar to Eq. \eqref{costlqr}
and Eq. \eqref{minJ}, but the cost starts from time-step $k$ instead. The optimal cost-to-go at time-step $k$ is a quadratic function given by
\begin{equation}\label{quadfun}
V(\delta{\bf x}_{k})=\frac{1}{2}\delta{\bf x}_k^{T}S_k\delta{\bf x}_k+\delta{\bf x}_k^{T}\mathbf{s}_k+ c_k,
\end{equation}
where $S_k$, $\mathbf{s}_k$, and $c_k$ are computed backwards in time from the final conditions $S_N=Q_N$, $\mathbf{s}_N=\mathbf{q}_N$, and $c_N=c$. Thus, the minimization of the control sequence becomes a minimization over a control input at a time-step which is known as the principle of optimality \cite{31}. To find the optimal value, the Ricatti equations are propagated backwards from the final conditions as given by
\begin{subequations}
\begin{equation}\label{rit1}
\begin{split}
S_k=A_k^{T}S_{k+1}A_k+Q_k-\left(B_k^{T}S_{k+1}A_k+P_k^{T} \right)^{T}\left(B_k^{T}S_{k+1}B_k+R_k \right)^{-1}\left(B_k^{T}S_{k+1}A_k+P_k^{T}\right),
\end{split}
\end{equation}
\begin{equation}\label{rit2}
\begin{gathered}
\mathbf{s}_k=\mathbf{q}_k+A_k^{T}\mathbf{s}_{k+1}+A_k^{T}S_{k+1}\mathbf{g}_k
\\-\left(B_k^{T}S_{k+1}A_k+P_k^{T}\right)^{T}\left(B_k^{\mathsf{T}}S_{k+1}B_k+R_k\right)^{-1}\left(B_k^{T}S_{k+1}\mathbf{g}_k+B_k^{T}\mathbf{s}_{k+1}+\mathbf{r}_k\right),
\end{gathered}
\end{equation}
\begin{equation}\label{rit3}
\begin{gathered}
c_k=\mathbf{g}_k^{T}S_{k+1}\mathbf{g}_k+2\mathbf{s}_{k+1}^{T}\mathbf{g}_k+c_{k+1}
\\-\left(B_k^{T}S_{k+1}\mathbf{g}_k+B_k^{T}\mathbf{s}_{k+1}+\mathbf{r}_k\right)^{T}\left(B_k^{\mathsf{T}}S_{k+1}B_k+R_k\right)^{-1}\left(B_k^{T}S_{k+1}\mathbf{g}_k+B_k^{T}\mathbf{s}_{k+1}+\mathbf{r}_k\right).
\end{gathered}
\end{equation}
\end{subequations}
Thus, approximately optimal LQR solutions from nonlinear equations of motion about a quadratic cost function can be found which can be used in conjunction with RRT* \cite{karaman2011sampling} to obtain dynamically feasible continuous trajectories with the asymptotic optimality property. LQR-RRT* consists of seven components including: 
\begin{itemize}
	\item \textit{Random sampling}: The state-space is randomly sampled uniformly to obtain a node (state $\mathbf{x}_{rand}$).
	\item \textit{Nearest node}: With $\mathbf{x}_{rand}$ and a current set of nodes $\mathbb{N}$ of the tree (trajectory), the nearest node in the tree relative to $\mathbf{x}_{rand}$ is obtained using the value function in Eq. \eqref{quadfun} given by
	\begin{equation}
	    \mathbf{x}_{nearest}=\argmin_{\mathbf{x}'\in \mathbb{N}}(\mathbf{x}'-\mathbf{x}_{rand})^TS_{rand}(\mathbf{x}'-\mathbf{x}_{rand})+(\mathbf{x}'-\mathbf{x}_{rand})\mathbf{s}_{rand}+c_{rand},
	\end{equation}
	where $S_{rand}$, $\mathbf{s}_{rand}$, and $c_{rand}$ are computed about the time-step which $\mathbf{x}_{rand}$ occurs.
	\item \textit{LQR steer}: An LQR trajectory is found between nodes $\mathbf{x}_{nearest}$ and $\mathbf{x}_{rand}$. Note that the trajectory found can be a path that moves towards $\mathbf{x}_{rand}$ with a final state $\mathbf{x}_{new}$.
	\item \textit{Near nodes}: Using $\mathbf{x}_{new}$ and set $\mathbb{N}$, a subset of nodes $\mathbb{N}_{near}\subseteq \mathbb{N}$ is found within the vicinity of $\mathbf{x}_{new}$ using Eq. \eqref{quadfun} given by
	\begin{equation}
	\resizebox{.9 \hsize}{!}{$ 
	     \left\{ \mathbf{x}'\in \mathbb{N}:(\mathbf{x}'-\mathbf{x}_{new})^TS_{new}(\mathbf{x}'-\mathbf{x}_{new})+(\mathbf{x}'-\mathbf{x}_{new})\mathbf{s}_{new}+c_{new}\leq\gamma\left(\frac{\text{log}n}{n}\right)^{1/n_{\mathbf{x}}}\right\}.$}
	\end{equation}
    \item \textit{Choosing a parent}: LQR trajectories for each node in $\mathbb{N}_{near}$ are found with respect to $\mathbf{x}_{new}$. The node with the lowest cost ($\mathbf{x}_{min}$) and the trajectory $\sigma_{min}$ becomes the parent node of $\mathbf{x}_{new}$. 
    \item \textit{Collision checking}: The trajectory $\sigma_{min}$ is checked against any obstacles within the state-space and is further discussed in the Collision Avoidance subsection.
	\item \textit{Rewire near nodes}: If $\sigma_{min}$ and $\mathbf{x}_{new}$ are collision-free, $\mathbf{x}_{new}$ is added to the set of nodes $\mathbb{N}$, and then attempts are made to reconnect $\mathbf{x}_{new}$ with the set $\mathbb{N}_{near}$ with LQR trajectories if the cost is less than its current parent node.
\end{itemize}
This algorithm provides a single pass of LQR-RRT* which generates trajectories built from sampling. This meets with on-orbit assembly goals like avoiding collisions with structural parts. Unfortunately, sampling through LQR-RRT* may provide jerky and unnatural paths if not enough samples are taken, thus, trajectory smoothing through LQR shortcutting is applied to reduce this effect and provide collision-free trajectories.

\subsection{Trajectory Smoothing by LQR Shortcutting}
Given the LQR-RRT* trajectory which may contain jerky and unnatural paths, trajectory smoothing can be directly applied using a shortcutting approach which iteratively constructs path segments between existing nodes \cite{kallmann2008planning,geraerts2007creating,hauser2010fast}. This enables the generation of high-quality, collision-free, smooth paths that can be used directly for control. Specifically, LQR shortcutting algorithm is presented which generates dynamically feasible shortcut segments that are optimal to the LQR cost function given by Eqs. \eqref{costlqr} and \eqref{immediateterminallqr}.
\begin{itemize}
\item \textit{Random sampling}: From the initial trajectory $\sigma_0$, two nodes, $\mathbf{x}_a$ and $\mathbf{x}_b$, are randomly sampled which results in a trajectory with three sections, $\sigma_0=\left[\sigma_1,\sigma_2,\sigma_3\right]$.
\item \textit{LQR Interpolation}: An interpolation using an LQR solution between $\mathbf{x}_a$ and $\mathbf{x}_b$ is determined. The new trajectory is $\sigma_{2,interp}$. This solution incorporates dynamic feasibility and optimality in the generated shortcut.
\item \textit{Collision Checking}: Lastly, the new trajectory $\sigma_{2,interp}$ is checked against any obstacles within the state-space which is discussed in the Collision Avoidance subsection. If the shortcut is collision-free, then $\sigma_{2,interp}$ is patched with the two other sections $\sigma_1$ and $\sigma_3$ into $\sigma_{new}=\left[\sigma_1,\sigma_{2,interp},\sigma_3\right]$.
\end{itemize}
By applying LQR shortcutting on the LQR-RRT* trajectory, high-quality, collision-free, smooth paths that are dynamically feasible for the on-orbit free-flyer is generated. Thus, control can be applied to track this trajectory.
\subsection{Collision Avoidance}
In order to prevent collisions between the free-flyer and the assembled structure, collision avoidance must be considered. As discussed throughout this section, LQR-RRT* and LQR shortcutting include a collision checking feature which is used to generate collision-free trajectories between two nodes. If the trajectory between two nodes intersects with an obstacle, the trajectory is thrown out, and a new trajectory is generated at the next iteration. The goal for the on-orbit free-flyer is to assemble a structure under inertial uncertainty while avoiding obstacle collisions. It is assumed that the obstacles themselves are tracked by either the free-flyer or another spacecraft. By knowing the obstacle states through obstacle tracking, the on-orbit free-flyer can approximate keep-out zones using a 3D ellipsoid constraint \cite{jewison2015model}.
\subsection{Ellipsoid Method}
The 3D ellipsoid can be used to create an individual keep-out zone for the obstacle in the state-space as well as consider uncertainty in the obstacle's size. For example, if there is no uncertainty in an obstacle's size, an ellipsoid can be formed to enclose the obstacle's volume. In the case where there is uncertainty in the obstacle's size, an ellipsoid can be formed which encloses the obstacle's volume with a safety factor. The 3D ellipsoid constraint is given by
\begin{equation}\label{ellcon}
    \left( \mathbf{x}_{pos}-\mathbf{x}_{obs}\right)^T P_{obs}\left( \mathbf{x}_{pos}-\mathbf{x}_{obs}\right)\geq 1,
\end{equation}
where $\mathbf{x}_{obs}$ is the ellipsoid's centroid position, $P_{obs}$ is a positive definite shape matrix of the ellipsoid, and $\mathbf{x}_{pos}$ is the position of a single state on the trajectory in question. 
The shape matrix, $P_{obs}$, can be designed for safety factors which considers the uncertainty in the obstacle's size. This constraint, Eq. \eqref{ellcon}, is nonlinear and nonconvex, but for sampling-based motion planning, it is used as a check for newly generated trajectory segments. Each state in the generated trajectory segment must be evaluated against Eq. \eqref{ellcon} for collisions. This is given by 
\begin{equation}\label{ellconfull}
    \begin{bmatrix}
          1 \\
           \vdots \\
           1
         \end{bmatrix}_{N\times 1} -\text{diag}\{ \mathcal{X}^T P_{obs}\mathcal{X}\}\leq \begin{bmatrix}
          0 \\
           \vdots \\
           0
         \end{bmatrix}_{N\times 1},
\end{equation}
where $N$ is the time horizon of the trajectory segment, $\mathcal{X}$ is a stacked sequence of position states specified by $\mathcal{X}=\left[ \left(\mathbf{x}_{pos,0}-\mathbf{x}_{obs,0}\right),\left(\mathbf{x}_{pos,1}-\mathbf{x}_{obs,1}\right),\dots,\left(\mathbf{x}_{pos,N}-\mathbf{x}_{obs,N}\right)\right]\in\mathbb{R}^{3\times N}$, and $\text{diag}(\cdot)$ transforms the diagonals of the $N\times N$ matrix into a column vector. Although this work assumes that obstacles are static once they are assembled, dynamic updates to the obstacle state, $\mathbf{x}_{obs}$, can also be integrated. The ellipsoidal obstacle constraint can be expanded for multiple obstacles in the state-space. As more 3D printed objects are assembled, the number of obstacles in the space increase. So each generated trajectory segment must be compared to every obstacle in the field as given by
\begin{equation}\label{ellconful2l}
    \begin{bmatrix}
          1 \\
           \vdots \\
           1
         \end{bmatrix}_{(N\cdot N_{obs})\times 1} -\begin{bmatrix}\text{diag}\{ \mathcal{X}^T P_{obs,1}\mathcal{X}\}\\
         \vdots \\
         \text{diag}\{ \mathcal{X}^T P_{obs,N_{obs}}\mathcal{X}\}
         \end{bmatrix}
         \leq \begin{bmatrix}
          0 \\
           \vdots \\
           0
         \end{bmatrix}_{(N\cdot N_{obs})\times 1},
\end{equation}
where $N_{obs}$ is the number of obstacles. With multiple obstacles, LQR-RRT* and LQR shortcutting can apply the 3D ellipsoidal constraint for collision avoidance.

Note, that considerations must be made for ill-posed motion planning problems. If the free-flyer moves with a high velocity, the time update for the position for the discrete system may jump quickly if the time interval is not fine enough. If the distance for two consecutive position states is larger than the smallest characteristic length of the ellipsoid, the algorithm will not detect a collision, although in reality (continuous time), a collision may occur. Methods to migitate these ill-posed problems is by increasing the sample resolution time and by interpolating the trajectory during the collision check \cite{jewison2017guidance}. So, carefully consideration is required to prevent an ill-posed problems for collision-free trajectory generation.

\section{Robust Tube Model Predictive Control}

Model predictive control (MPC) repeatedly solves an online finite horizon optimal control problem for general dynamical systems under constraints. At each time step, the first output is executed and the computation begins again with any necessary system updates. While proofs for MPC stability under certain conditions \cite{Mayne2011} exist, guarantees for stochastic systems are generally lacking. Robust tube MPC is a notable exception, providing robustness guarantees when bounded additive uncertainty is encountered in the system dynamics. Using tube MPC, a portion of control authority is reserved for robust actuation, often in a simple feedback form to counter disturbances. The guarantee obtained is one of tube robustness---if a system starts in a tube of possible states, it remains within a tube around a nominal trajectory, within a set termed the robust positively invariant set (RPI). These tubes are formed around the nominally planned MPC trajectory, and the motion of the system can be thought of as a composition of these tubes as the control trajectory is replanned online \cite{Mayne2005}.

\subsection{Controller Formulation} Linear robust tube MPC is now introduced in order to meet the requirements of a robust control technique for the problem formulation. A discrete-time LTI system has dynamics 

\begin{align} \begin{split}
\mathbf{x}_{k+1} &= \mathbf{A}\mathbf{x}_k + \mathbf{B}\mathbf{u}_k + \mathbf{w}_k\\
\mathbf{x} &\in \mathcal{X} \subseteq \mathbb{R}^{n_\mathbf{x}}\\
\mathbf{u} &\in \mathcal{U} \subseteq \mathbb{R}^{n_\mathbf{u}}\\
\mathbf{w} &\in \mathcal{W} \subseteq \mathbb{R}^{n_\mathbf{u}}
\end{split}, \end{align}

where $\mathcal{W}$ is a \textit{convex, bounded, compact disturbance}. Modeling $\mathbf{w}_k$ is a matter of defining a stepwise bounded additive uncertainty that adequately captures system uncertainty at each timestep. $\mathbf{w}_k$ then can be used to model parameter uncertainty, external disturbances, or unmodeled dynamics. $\mathcal{W}$ is very frequently assumed to be a polytopic set to simplify computation.

Robust positively invariant (RPI) sets for tube MPC are those which are unchanged given a disturbance sequence $\mathbf{w}_{0:k}$ and are generally referred to as $\mathcal{Z}$ (used interchangeably here with the approximation of this set). $\mathcal{Z}$ is RPI if for any initial state $\mathbf{x}_0 \in \mathcal{Z}$ and all disturbance sequences $\mathbf{w}_{0:k} \in \mathcal{W}$ for $k > 0$, then $\mathbf{x}_k \in \mathcal{Z}$. $\mathcal{Z}^*$ is the minimal RPI (mRPI), the smallest possible RPI which is often only approximately known and is often a polytopic approximation. Robust tube methods are based on the idea that the combination of all possible uncertainty realizations applied to the controller results in a bundle of trajectories relative to the nominal system, where $\mathbf{w}_{0:k} = \mathbf{0}$. The nominal system is then denoted as

\begin{align}
\mathbf{z}_{k+1} = \mathbf{A}\mathbf{z}_k + \mathbf{B}\mathbf{v}_k.
\end{align}

First, the nominal system inputs are calculated using direct shooting MPC with a free (but penalized) initial state. This produces a nominal trajectory, $\mathbf{z}_{0:k}, \mathbf{v}_{0:k}$, often abbreviated as $\mathbf{z}, \mathbf{v}$. Define the receding horizon MPC problem as 

\begin{align}
	\mathbf{z}, \mathbf{v} \leftarrow \mathcal{M}(\mathbf{x}, \mathbf{x}_{des}, \mathbf{u}_{des}, \mathcal{C}),
\end{align}

where $\mathcal{C}$ represents a set of modified constraints on the nominal MPC. The output of this procedure is the nominal input and state histories, with the notable modification from standard MPC that $\mathbf{z}_0$ is not constrained to match the real system state.

Secondly, a disturbance rejection controller operates on top of this nominal trajectory obeying control law

\begin{align}
\mathbf{u}^{dr}_k = \mathbf{K}(\mathbf{x}_k - \mathbf{z}_k).
\end{align}

This controller provides disturbance rejection, and the choice of gain $\mathbf{K}$ modifies the shape of the RPI set $\mathcal{Z}$ around $\mathbf{z}$. The control inputs are combined at each time step

\begin{align}
\mathbf{u}_k = \mathbf{v}_k + \mathbf{u}^{dr}_k.
\end{align}

In order to determine the nominal trajectory, additional constraints $\mathcal{C}$ must be added to the MPC to ensure constraint satisfaction when both the nominal and disturbance-rejection controllers are in use.

\subsection{RPI Set Calculation}

The deviation between the real system state, $\mathbf{x}_k$, and the planned feedforward nominal system state is

\begin{align}
\mathbf{e}_k \triangleq \mathbf{x}_k - \mathbf{z}_k.
\end{align}

This means the closed-loop error dynamics are

\begin{align}\begin{split}
\mathbf{e}_{k+1} &= (\mathbf{A} + \mathbf{B}\mathbf{K}) \mathbf{e}_k + \mathbf{w}_k\\
\mathbf{A}_{CL} &\triangleq \mathbf{A} + \mathbf{B}\mathbf{K}\\
\mathbf{e}_{k+1} &= \mathbf{A}_{CL} \mathbf{e}_k + \mathbf{w}_k
\end{split},\end{align}

where $\mathbf{A}_{CL}$ must be stable (Hurwitz). An RPI exists for the error dynamics\footnote{Note that $\oplus$ indicates the Minkowski sum.} \cite{Kolmanovsky1998},

\begin{align}
\mathbf{A}_{CL} \mathcal{Z} \oplus \mathcal{W} \subseteq \mathcal{Z}.
\end{align}

This means that if a state begins within $\mathcal{Z}$ and evolves under the given controller, it will remain within $\mathcal{Z}$ no matter what $\mathcal{W}$ is applied. If the nominal system trajectory $\mathbf{z}, \mathbf{v}$ can be found under tightened constraints then constraint satisfaction using the disturbance rejection control against that nominal trajectory is guaranteed. How one finds an approximate $\mathcal{Z}$ and $\mathbf{K}$ are one of the design challenges of robust tube MPC---many techniques exist including Limon's tuning procedure \cite{Limon2008}. $\mathbf{K}$ is frequently chosen as the finite horizon discrete LQR gain, which is used here.

The RPI set can be approximated using a polytope efficiently, which is necessary in forming constraints on the nominal MPC. First, $\mathbf{K}$, the disturbance rejection gain, is selected. The set is computed using the Minkowski sum of possible system evolutions under system perturbation for iterations $s \rightarrow \infty$,

\begin{align}
	\mathcal{Z}_s = \bigoplus_{j=0}^s (\mathbf{A}+\mathbf{B}\mathbf{K})^j \mathcal{W}.
\end{align}

Each iteration grows the set using the closed-loop perturbed system evolution. $s$ iterations may be used, up to a desired limiting $\epsilon$. Algorithms have been developed to approximate $\mathcal{Z}$ with a polytopic set \cite{Mayne2005,Rakovic2008a}, which this work follows.

\subsection{Standard MPC}

A standard MPC is set up following the usual MPC formulation. A terminal cost is defined relative to a desired reference trajectory, $\mathbf{x}_{des}$

\begin{align}
	V_f = [\mathbf{z}_N-\mathbf{x}_{N,des}]^\top\mathbf{H}[\mathbf{z}_N-\mathbf{x}_{N,des}],
\end{align}

and serves as a control Lyapunov function, providing stability guarantees via

\begin{align}
V_f(\mathbf{z}_{k+1}) + l(\mathbf{z}_k, \mathbf{v}_k) \leq V_f(\mathbf{z}_k),
\end{align}

where $l(\mathbf{z}_k, \mathbf{v}_k)$ is the stage cost. The full nominal MPC $\mathcal{M}$ is
\begin{equation}
\begin{gathered}
 J=  \frac{1}{2}\sum_{k=0}^{N-1}[\mathbf{z}_k-\mathbf{x}_{k,des}]^\top\mathbf{Q}[\mathbf{z}_k-\mathbf{x}_{k,des}] + \frac{1}{2}[\mathbf{v}_k-\mathbf{u}_{k,des}]^\top\mathbf{R}[\mathbf{v}_k-\mathbf{u}_{k,des}]
 \\+\frac{1}{2}[\mathbf{z}_N-\mathbf{x}_{N,des}]^\top\mathbf{H}[\mathbf{z}_N-\mathbf{x}_{N,des}]\\
  \begin{aligned}
 \text{subject to}\hspace{10pt}& 
\mathbf{x}_0 = \mathbf{x}_0\\
&\mathbf{x}_{k+1} = f(\mathbf{x}_k,\mathbf{u}_k)\\        
&\mathbf{u}_{min} \leq \mathbf{u} \leq \mathbf{u}_{max}\\ 
&\mathbf{x} \in \mathcal{X}_{free}
\end{aligned}
\end{gathered},
\end{equation}

\subsection{Robust MPC Controller}

The standard MPC must be modified to guarantee constraint satisfaction while operating alongside the disturbance rejection controller in the robust MPC case. A few changes are needed: (1) the initial state of the nominal system does not need to be the same as the real system (i.e. the nominal initial state just needs to ``lasso" the real state within $\mathcal{Z}$); (2) the disturbance rejection controller is deployed on top of the nominal inputs $\mathbf{v}$; (3) the nominal robust MPC has tightened constraints. The nominal robust MPC problem can be written as
\begin{equation}
\begin{aligned}
&\underset{\mathbf{v},\bar{\mathbf{z}}_0, \bar{\bm{\theta}}}{\text{minimum}}\ &&  J= \sum_{k=0}^{N-1}[\bar{\mathbf{z}}_k-\mathbf{x}_{k,des}]^\top\mathbf{Q}[\bar{\mathbf{x}}_k-\mathbf{x}_{k,des}] + [\bar{\mathbf{v}}_k-\mathbf{u}_{k,des}]^\top\mathbf{R}[\bar{\mathbf{v}}_k-\mathbf{u}_{k,des}]\\
&&&\hphantom{J}+ [\bar{\mathbf{z}}_N-\mathbf{x}_{N,des}]^\top\mathbf{H}[\bar{\mathbf{z}}_N-\mathbf{x}_{N,des}]+
[\bar{\bm{\theta}}-\bm{\theta}_{des}]^\top\mathbf{T}[\bar{\bm{\theta}}-\bm{\theta}_{des}]\\
&\text{subject to}                                  &&\mathbf{z}_{k+1} = f(\mathbf{z}_{k},\mathbf{v}_{k})\\
&&& {\mathbf{z}} \in \bar{\mathcal{X}}\\        
&&& {\mathbf{v}} \in \bar{\mathcal{U}}\\ 
&&& ({\mathbf{z}}_N, \bar{\bm{\theta}}) \in \Omega^e\\
&\text{where}
&&\bar{\mathcal{X}} \triangleq \mathcal{X} \ominus \mathcal{Z}\\
&&&\bar{\mathcal{U}} \triangleq \mathcal{U} \ominus \mathbf{K}\mathcal{Z}\\
&&&\mathbf{z}^e \triangleq (\mathbf{z}, \theta)
\end{aligned}.
\end{equation}

$\mathbf{z}^e$ is the extended state and $\Omega^e$ is an augmented set containing the state RPI and a set parameterizing the terminal set $\bm{\theta}$. In the problem presented here, the $\bm{\theta}$ parameterization is dropped. The solution $\mathbf{v}$ is added to the disturbance rejection controller solution, $\mathbf{u}^{dr}$ to produce the control at each time step.

\subsection{Implications for On-Orbit Assembly Robustness}

The significance of the above design procedure is that arbitrary disturbances within $\mathcal{W}$ can be accounted for in the design procedure and a guarantee on system evolution within a robust tube can be provided. This allows one to use online, adaptable model predictive control with an explicit guarantee on system safety, which is especially critical when operating near fragile space structures on-orbit. However, one still needs to define the uncertainty set.

For this specific problem formulation, uncertainty is assumed to enter only through parametric mass uncertainty, meaning that the mass statistics, $\mathcal{N}(\mu_m, \sigma_m)$ must somehow be converted into stepwise uncertainty bounds on the discrete dynamics. In this case, the 2-$\sigma$ mass bounds of the latest available estimate are propagated through the discrete-time dynamics at the maximum system inputs and compared to the result using $\mu_m$ which is given by

\begin{align}
    \mathbf{w}_{max} &= |f(\mathbf{u}_{max}, m_0) - \texttt{max}\left( f(\mathbf{u}_{max}, m_{2 \sigma}), f(\mathbf{u}_{max}, m_{-2 \sigma}) \right) |.
    \label{eqn:approx}
\end{align}

Eq. \ref{eqn:approx} results in an approximation of the stepwise uncertainty bound under a  threshold of possible mass error values. This threshold is informed by the certainty in the mass estimate itself through its variance; higher variance results in a larger bound and greater conservativeness. In general, conversion of parametric uncertainty to a bounded stepwise uncertainty through a similar procedure allows tube robustness guarantees to be applied with a desired level of confidence.

\section{Information Rich Trajectory Design and Parameter Estimation}

There are many possible ways to produce information-rich trajectories, often called excitation trajectories. Approaches include E2-log maximization for increasing observability, POMDP formulations for parameter learning, and other information-theoretic weighting methods \cite{Swevers1997,Webb2014,wilson2014trajectory}. The goal of this trajectory design is to move an unknown system in such a way that the maximal amount of information is provided to an estimation scheme. In the context of on-orbit assembly of rigid objects, the parameters of interest are $\theta = \left\{m, \mathbf{x}_{cm}, I_{zz}, I_{yy}, I_{xx}, \mathcal{F}\right\}$, where $\mathcal{F}$ is the principal axes pose offset. An optimal control problem formulation can be used, where the cost function $J$ can be a metric such as Fisher information conditioned on these parameters. Fisher information is essentially a weighting on how much information content measurements have of system unknowns, written as

\begin{equation} \label{eq:FIM}
    F = E \left\{ \left[\frac{\partial}{\partial\pmb{\theta}}\ln{[p(\Tilde{\mathbf{y}}|\pmb{\theta}})] \right] \left[\frac{\partial}{\partial\pmb{\theta}}\ln{[p(\Tilde{\mathbf{y}}|\pmb{\theta}})] \right] ^T\right\}.
\end{equation}.

The information rich trajectory must be planned by taking into account the actuation, state and space constraints. The following optimization problem is solved for N time steps given by
 
\begin{equation}\label{opt}
\begin{aligned}
 &\underset{\mathbf{x, u}}{\text{minimize}} && tr\left(F ^{-1}\right)\\
 & \text{subject to}
& & \mathbf{x}_{k+1} = f(\mathbf{x}_{k},\mathbf{u}_{k}), k = 0,...,N \\
&&& \mathbf{u}_{min} \leq \mathbf{u}_{k} \leq \mathbf{u}_{max},k = 0,...,N  \\
&&& \mathbf{x}_{0} = \mathbf{x}_{start}\\
&&& \mathbf{x}_{N} = \mathbf{x}_{des} \\
&&& \mathbf{x}_{k} \in \mathcal{X}_{free}, k = 0,...,N\\
\end{aligned},
\end{equation}
 where ${X}_{free}$ is the obstacle-free space available for robot system identification. $F = \sum_{k = 1}^{N}{F_k}$ denotes the total value of Fisher information over the trajectory. The result of this optimal control procedure is a trajectory $\mathbf{x}_{opt}(k)$
 which is optimally exciting in terms of some weighting of the Fisher information matrix over the parameters $\theta$ (in this work, the trace of the inverse of the FIM, also known as the A-optimality criterion, is minimized). 
 
 \section{Estimation scheme}
  Applying an estimation scheme to the actual input and state histories acquired after tracking the exciting trajectory allows for updating of $\theta$. The estimated parameters may then be handed off to motion planning and control to create a more accurate system model.

 In this work, a Maximum Likelihood batch-estimation method \cite{crassidis2011} is used for parameter estimation, and all data from an information-rich segment of the trajectory is be processed at once. Considering that pose estimates are available from a localization algorithm and the control inputs are known, the estimator solves the following optimization problem given by \cite{burri2018framework, kaess2008isam}

\begin{equation}
\hat{\mathbf{x}},\hat{\mathbf{\theta}} = \argmaxF_{\mathbf{x}, \mathbf{\theta}} \prod_{k = 1}^n p(\tilde{y}_k|\mathbf{x}_k)\prod_{k = 1}^np(\mathbf{x}_k|\mathbf{x}_{k-1},\mathbf{u}_{k-1}, \mathbf{\theta}) .
\end{equation}

The measurement and dynamic models are linearized by perturbing about a nominal trajectory, $\mathbf{x} = \mathbf{\overline{\mathbf{x}}} \boxplus \delta{\mathbf{x}}$,  $\theta = \mathbf{\overline{\mathbf{\theta}}} + \delta{\mathbf{\theta}}$. Here $\boxplus$ signifies rotation composition for the quaternions and addition for all other states. Assuming the noise to be Gaussian, the estimation problem ultimately reduces to
\begin{equation}
    \delta{\pmb{\vartheta}} = \underset{\delta{\pmb{\vartheta}}}{\text{minimize}}||b(\overline{\pmb{\vartheta}}) + A\delta{\pmb{\vartheta}}||^2, 
\end{equation}
where  $\pmb{\vartheta} = \left[\mathbf{x}_n^T,\mathbf{x}_{n-1}^T,...,\pmb{\theta}^T \right]^T$ denotes the vector of parameters to be estimated. The nominal residuals $\tilde{y}_i \boxminus h(\overline{x}_k)$ and $x_{k+1} - f(\overline{x}_k, \overline{\theta},\overline{u}_k)$ are grouped under $b(\overline{\pmb{\vartheta}}$), while $A$ is a function of the Jacobians $\frac{\partial  h(\overline{x}_k)}{\partial x_k}$,  $\frac{\partial  f(\overline{x}_k, \overline{\theta}_k, u_k)}{\partial {x}_k}$, $\frac{\partial  f(\overline{x}_k, \overline{\theta}_k, u_k)}{\partial \theta_i}$ and $\frac{\partial  x_{k+1}}{\partial x_k}$. A few iterations of the estimation problem are performed and the nominal values are updates after each iteration as $  \mathbf{\overline{\mathbf{x}}} \boxplus \delta{\mathbf{x}}$ and $\mathbf{\overline{\mathbf{\theta}}} + \delta{\mathbf{\theta}}$. The estimate covariance for the inertial parameters can be extracted from $(A^TA)^{-1}$ and converted to uncertainty bounds for use by the robust MPC.
 
\section{Robotic Free-Flyer Dynamics}
A double integrator and rigid body system are used to describe the free-flyer dynamics for on-orbit assembly of space structures. In this work, it is assumed that effects due to relative motion is negligible. The on-orbit free-flyer and the 3D printed structural object are within a close vicinity of each other. The states of the on-orbit free-flyer are given by
\begin{equation}
        \mathbf{x}(t)=\begin{bmatrix}
          \mathbf{r}(t) \\
          \mathbf{v}(t) \\
          \mathbf{q}(t) \\
          \boldsymbol{\omega}(t)
         \end{bmatrix},\hspace{12pt}
         \begin{matrix}
          \mathbf{r}(t)=\left[ r_x(t),\; r_y(t),\; r_z(t)\right]^T \\
          \mathbf{v}(t)=\left[ v_x(t),\; v_y(t),\; v_z(t)\right]^T  \\
          \mathbf{q}(t) = \left[ q_x(t),\; q_y(t),\; q_z(t),\; q_s(t)\right]^T \\
          \boldsymbol{\omega}(t)=\left[ \omega_x(t),\; \omega_y(t),\; \omega_z(t)\right]^T 
         \end{matrix},
\end{equation}
where $\mathbf{r}(t)$, $\mathbf{v}(t)$, $\boldsymbol{q}(t)$, and $\boldsymbol{\omega}(t)$ are the position, velocity, orientation, and angular velocity of the rigid body, respectively. For translational motion, $\mathbf{r}(t)$ and $\mathbf{v}(t)$ can be propagated by double integrator dynamics about the rigid body center of mass (CoM) given by
\begin{subequations}\label{doubleintdynamics}
\begin{equation}
    \mathbf{\dot{r}}_{CoM}(t)=\mathbf{v}_{CoM}(t),
\end{equation}
\begin{equation}
    \mathbf{\dot{v}}_{CoM}(t)=\frac{\mathbf{f}(t)}{m},
\end{equation}
\end{subequations}
where $m$ and $\mathbf{f}(t)$ is the mass and the force exerted by the free-flyer, respectively. The rigid body attitude dynamics which propagates $\boldsymbol{q}(t)$ and $\boldsymbol{\omega}(t)$ is given by
\begin{subequations}\label{rigiddynamics}
\begin{equation}
    \mathbf{\dot{q}}(t)=\frac{1}{2}\Xi(\mathbf{q}(t))\boldsymbol{\omega}(t),
\end{equation}
\begin{equation}
    \boldsymbol{\dot{\omega}}(t)=-\mathbf{I}^{-1}\boldsymbol{\omega}(t)\times\mathbf{I}\boldsymbol{\omega}(t)+\mathbf{I}^{-1}\boldsymbol{\tau}(t)
\end{equation}
\end{subequations}
where $\mathbf{I}$ is the moment of inertia, $\boldsymbol{\tau}$ is the torque exerted by the free-flyer, and $\Xi(\mathbf{q}(t))$ is
\begin{equation}
    \Xi(\mathbf{q}(t))=\begin{bmatrix}
          q_s(t)&-q_z(t)&q_y(t)\\
          q_z(t)&q_s(t)&-q_x(t) \\
          -q_y(t) &q_x(t)& q_s(t) \\
          -q_x(t)&-q_y(t)&-q_z(t)
         \end{bmatrix}.
\end{equation}
The dynamics given by Eqs. \eqref{doubleintdynamics} and \eqref{rigiddynamics} can be formed as Eq. \eqref{fxu} where $\mathbf{u}(t)=\left[\mathbf{f}(t),\; \boldsymbol{\tau}(t)\right]^T$ and can be linearized and discretized for motion planning and control given by Eq. \eqref{axbu}.
\section{Results}
For this work, the Astrobee on-orbit free-flyer is tasked for on-orbit assembly of space structures using the architecture in Figure \ref{planblockdiagram} which involves the motion planning to the 3D printer (in which parts are manufactured), to the safe area for estimation of unknown inertial parameters, and finally to the assembly area for construction. Trajectories are planned using LQR-RRT* and LQR shortcutting in Segments 1 and 3 while rich trajectory generation is used for inertial estimation in Segment 2. The inertial parameters of the Astrobee is known while the Astrobee with object inertial parameters are  given by Table \ref{tab:AIP}. These inertial parameters are considered truth for this simulation. For control, the first segment uses standard MPC since the Astrobee inertial properties are known, but for the second and third segment, the Astrobee with object inertial properties are uncertain, thus robust tube MPC is used to mitigate the effects of inertial uncertainty. Lastly, it is assumed that the attitude state is perfectly tracked using standard and robust tube MPC. Attitude tracking is an area of future work.

 \begin{table}[htbp]
   \caption{Astrobee: Inertial Properties} \label{tab:AIP}
   \small 
   \centering 
   \begin{tabular}{lcccc} 
   \toprule[\heavyrulewidth]\toprule[\heavyrulewidth]
   \textbf{Properties} & \textbf{Mass [kg]}&\multicolumn{3}{c}{\textbf{Moment of Inertia [kg m$^\text{2}$]}} \\ 
    & & \textbf{I$_{\text{xx}}$} &\textbf{I$_{\text{yy}}$}&\textbf{I$_{\text{zz}}$}\\
   \midrule
   Astrobee   & 9.0877&0.1454&0.1366&0.1594 \\
   Astrobee with object & 15 &0.1464&0.1376&0.1604\\
   \bottomrule[\heavyrulewidth] 
   \end{tabular}
\end{table}

\begin{figure}[!htb]
\begin{centering}
    \subfigure[Position time history]{
\includegraphics[keepaspectratio,trim={2.35cm 1.05cm .85cm .1cm},clip,width=.48\textwidth]{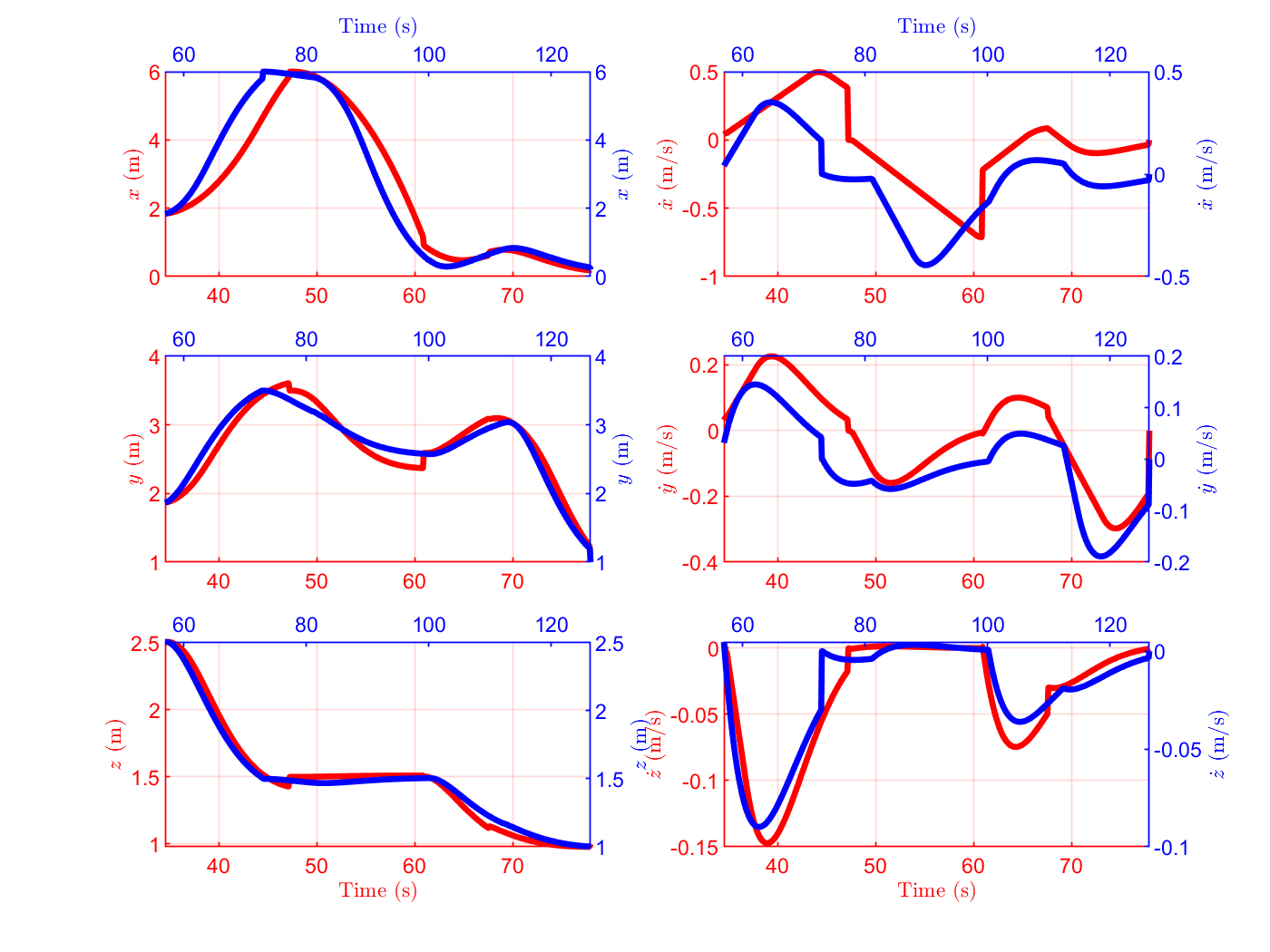}
      \label{fig:rrtsmoothpos}}
\subfigure[Attitude time history]{
\includegraphics[keepaspectratio,trim={2.35cm 1.05cm .85cm .1cm},clip,width=.48\textwidth]{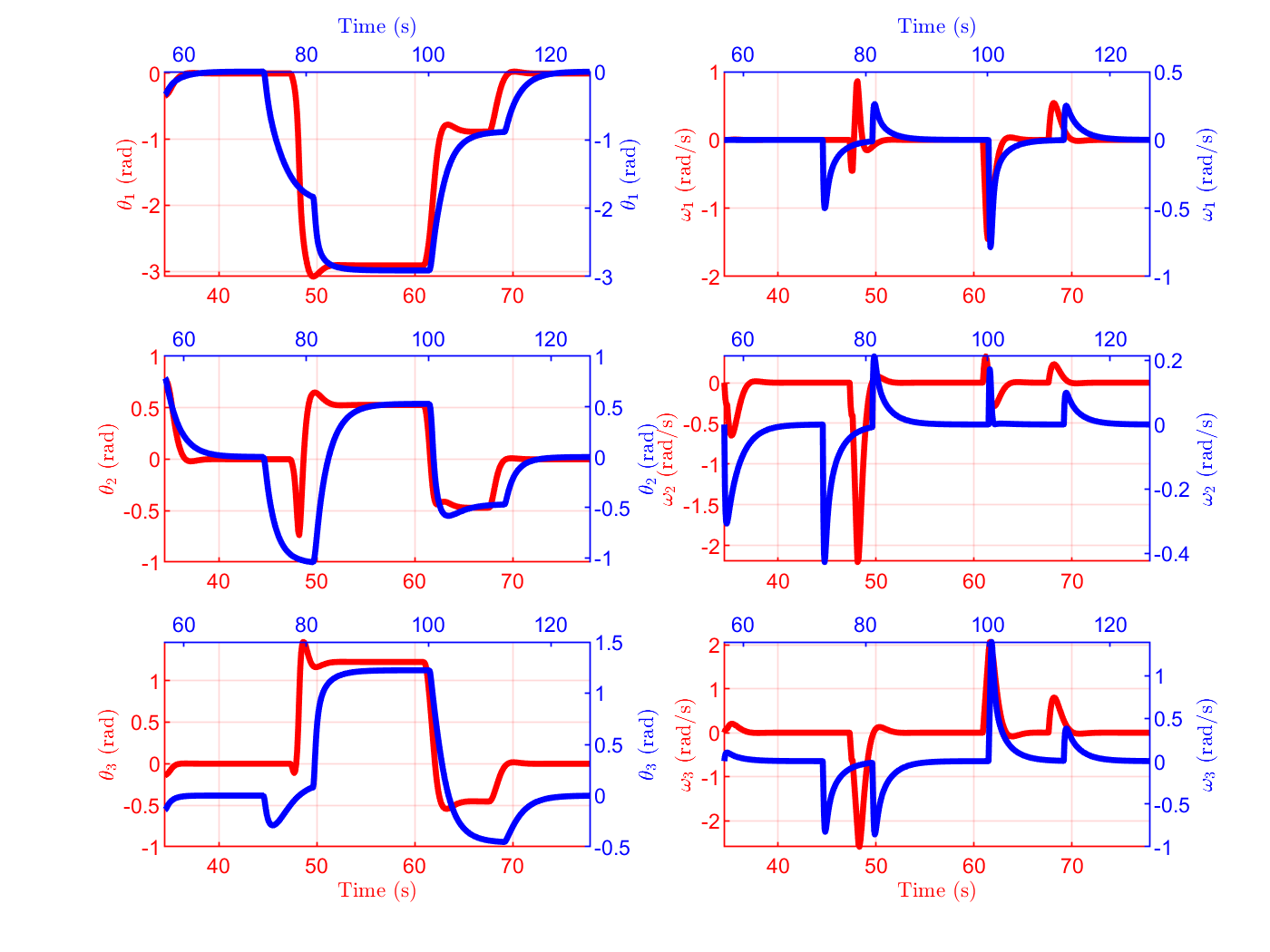}
    \label{fig:rrtsmoothatt}} 
        \caption{Trajectories for the Astrobee free-flyer using LQR-RRT* (blue) and LQR shortcutting (red) for segments 1 and 3.}\label{fig:rrtsmooth}
\end{centering}
\end{figure}

From the simulation of first and third segment, Figure \ref{fig:rrtsmooth} shows the trajectories obtained from LQR-RRT* and LQR shortcutting given by the blue and red lines, respectively. At specific points along the path, the LQR-RRT* trajectory is jerky and unnatural since the algorithm is run until an initial trajectory, which completes the path between Astrobee and the desired state, is found. If the LQR-RRT* algorithm continues sampling the space through time, the trajectories obtained would reach asymptotic optimality with respect to the cost function, but for this work, computational efficiency is necessary for Astrobee to perform on-orbit assembly. With this jerky and unnatural trajectory LQR shortcutted trajectory is found which is able to get to the target in a shorter amount of time by taking shortcuts. Thus, it reduces the effects of the randomness that occurs from sampling using LQR-RRT*. Thus, for the first and third segment, LQR shortcutting is applied directly after a LQR-RRT* trajectory is found to provide smooth collision-free trajectories.

\begin{figure}[!htb]
\begin{centering}
    \subfigure[Segment 1: Sampling-based plan (red) and standard MPC (green)]{
\includegraphics[keepaspectratio,trim={2.35cm 1.75cm 3.1cm 1.75cm},clip,width=.48\textwidth]{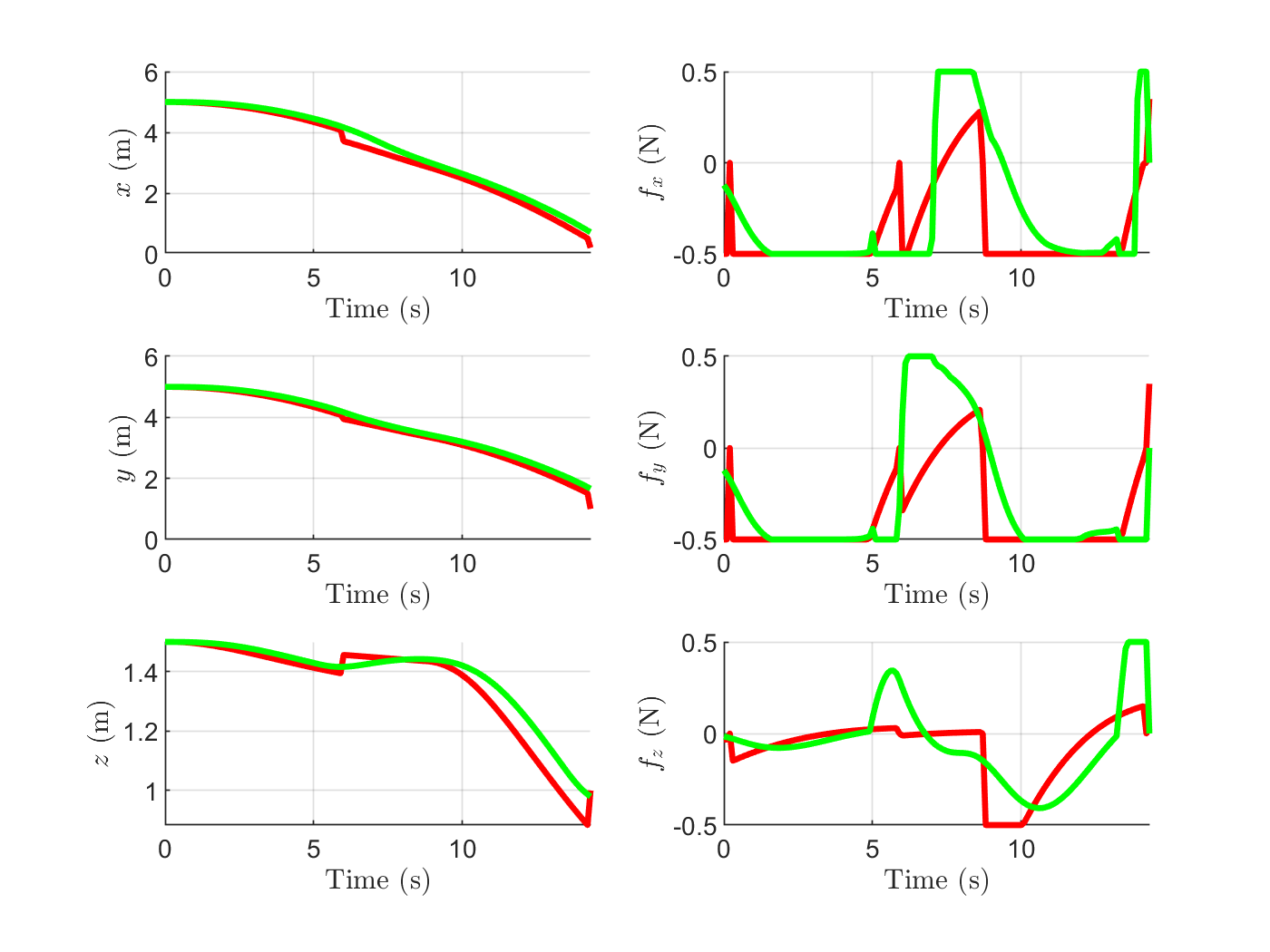}
      \label{fig:mpcseg1}}
\subfigure[Segment 2: Info. rich plan (red) and robust MPC (green) ]{
\includegraphics[keepaspectratio,trim={2.35cm 1.75cm 3.1cm 1.75cm},clip,width=.48\textwidth]{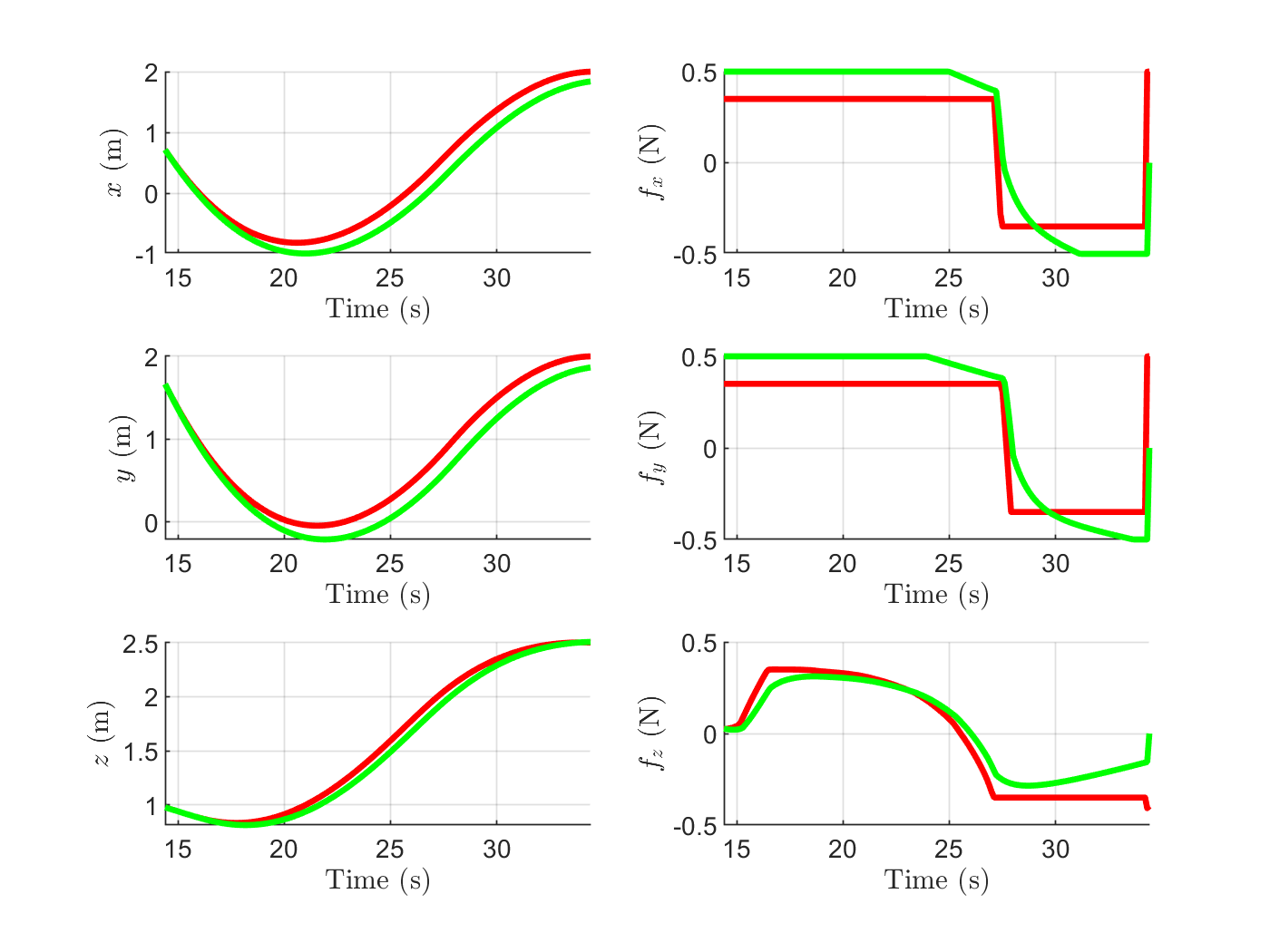}
    \label{fig:mpcseg2}} 
    \subfigure[Segment 2: Inertia Estimate (red) with covariance (black) and truth (blue) ]{
\includegraphics[keepaspectratio,trim={2.35cm 1.75cm 3.1cm 1.75cm},clip,width=.48\textwidth]{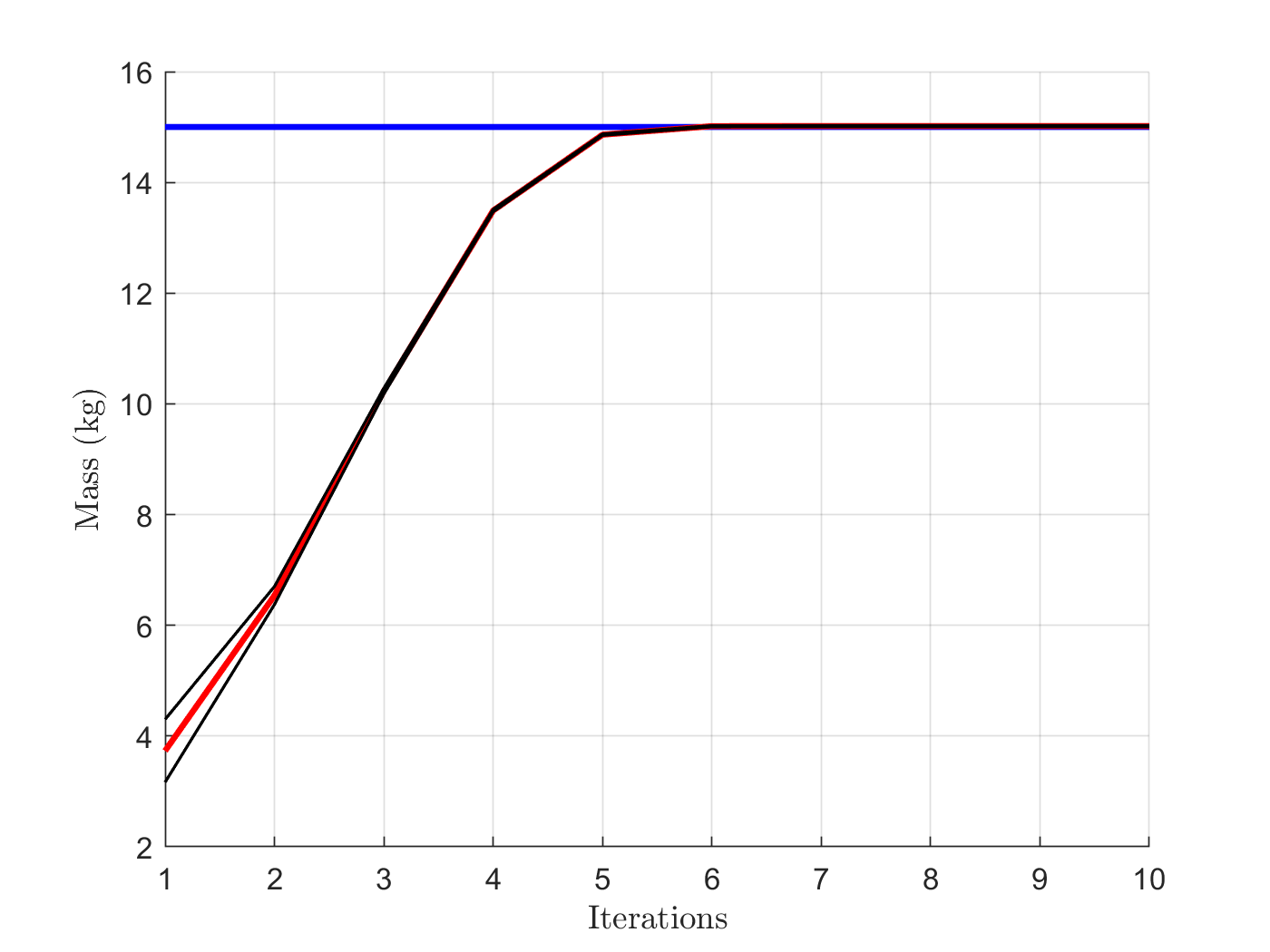}
    \label{fig:inertiaseg2}} 
\subfigure[Segment 3: Sampling-based plan (red), robust MPC - estimates (green), robust MPC - no estimates (magenta), and standard MPC, no estimates (cyan)]{
\includegraphics[keepaspectratio,trim={2.1cm .90cm 3.0cm 1.75cm},clip,width=.48\textwidth]{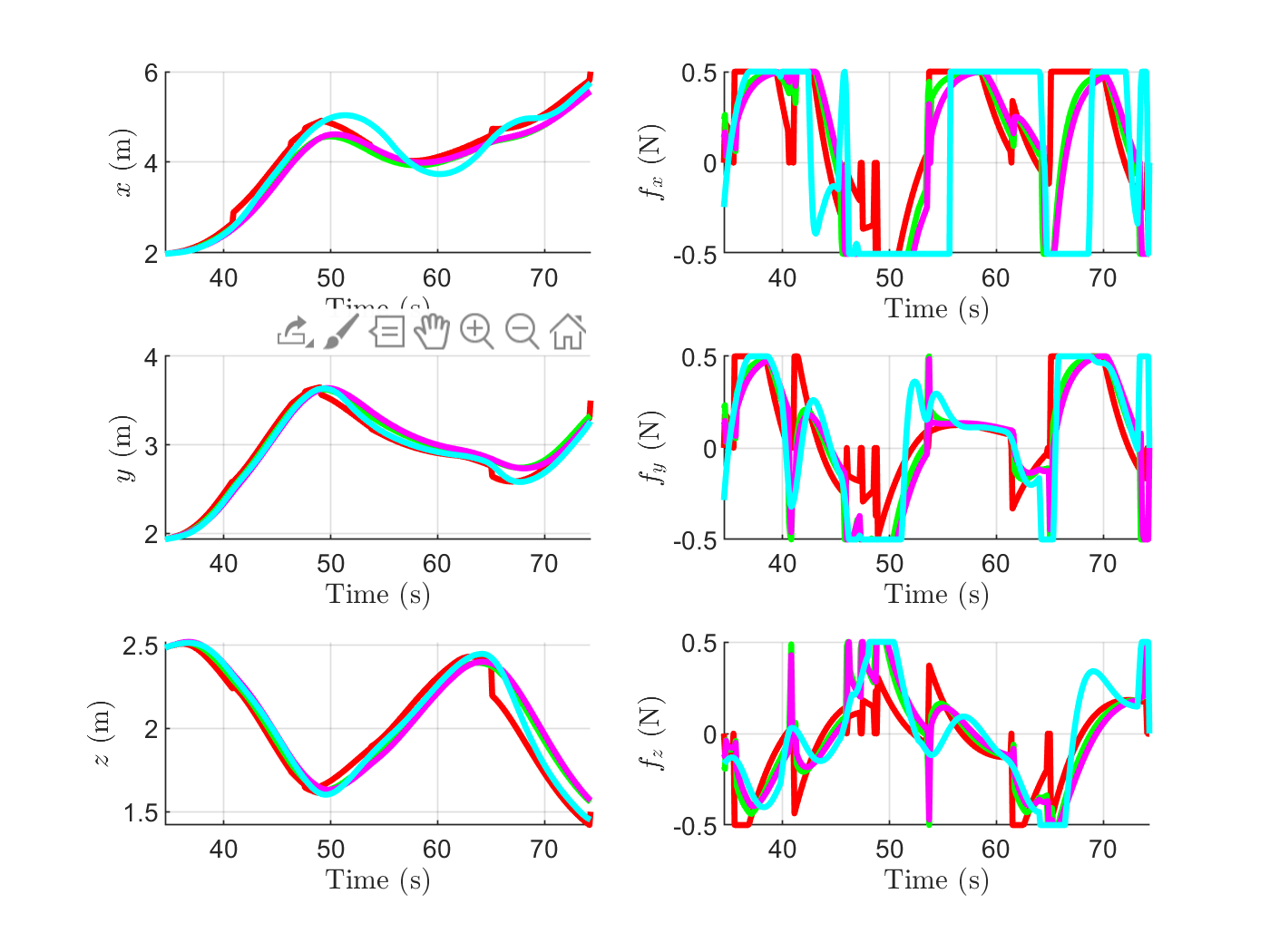}
    \label{fig:mpcseg3}} 
        \caption{Planner, control, and estimation responses of the Astrobee free-flyer for segments 1, 2, and 3.}\label{fig:mpc}
\end{centering}
\end{figure}

\begin{figure}[!htb]
\begin{centering}
    \subfigure[Segment 1: Motion from the assembly area to the 3D printer]{
\includegraphics[keepaspectratio,trim={0cm .00cm 0cm .0cm},clip,width=.48\textwidth]{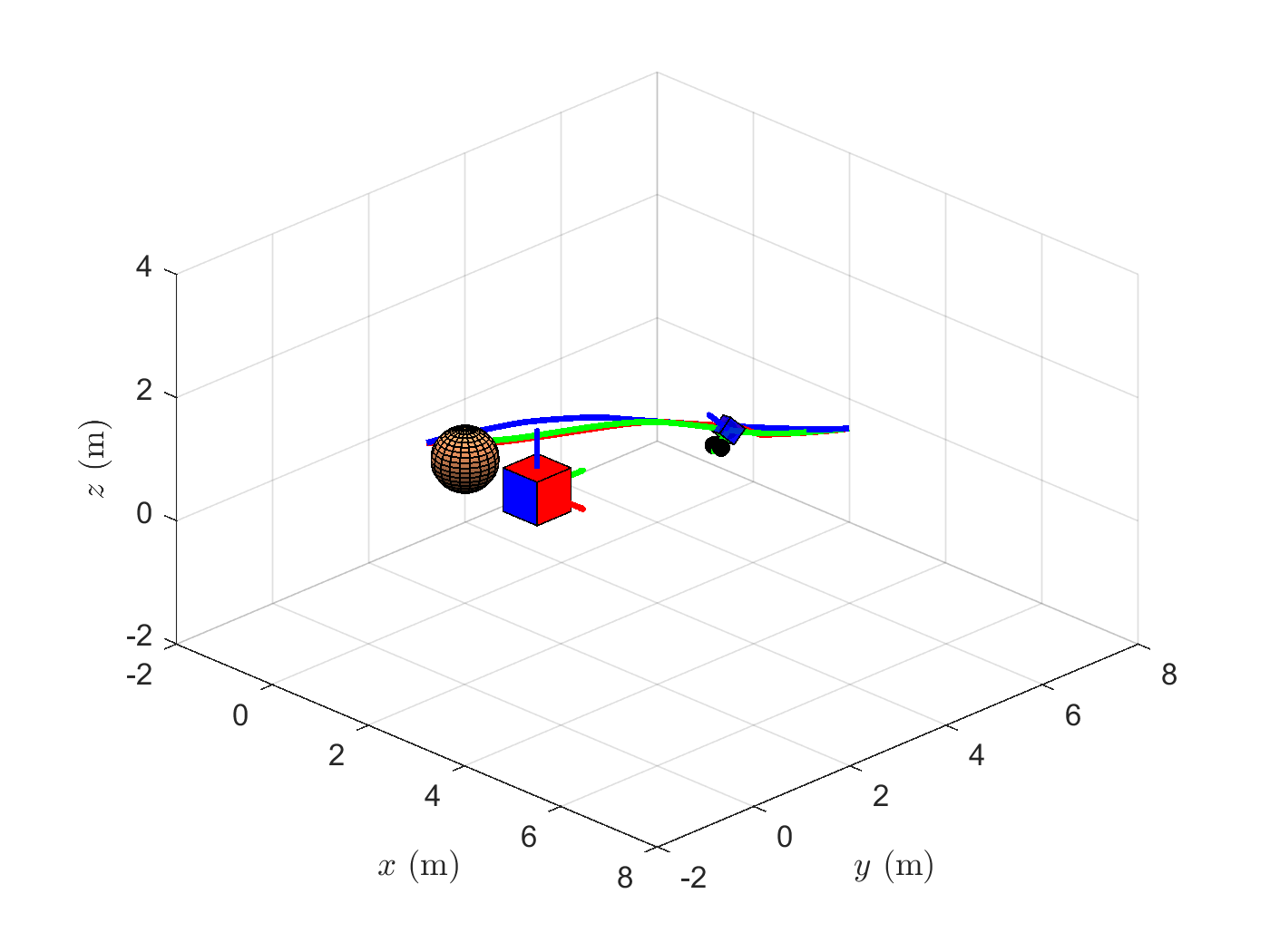}
      \label{fig:aa1}}
\subfigure[Segment 2: Motion from the 3D printer to the safe area]{
\includegraphics[keepaspectratio,trim={0cm .00cm 0cm .0cm},clip,width=.48\textwidth]{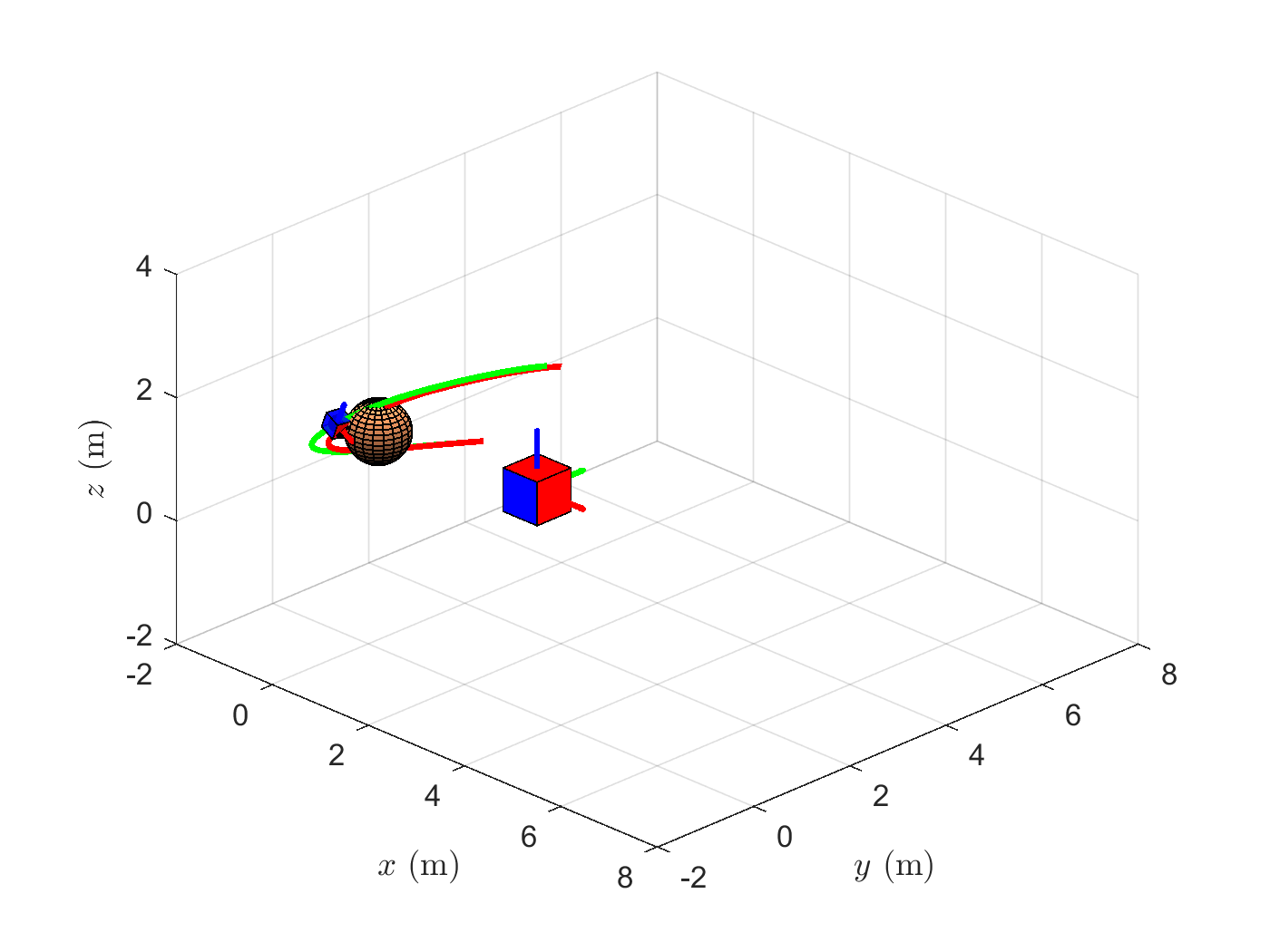}
    \label{fig:aa2}} 
\subfigure[Segment 3: Motion from the safe area to the assembly area]{
\includegraphics[keepaspectratio,trim={0cm .00cm 0cm .0cm},clip,width=.48\textwidth]{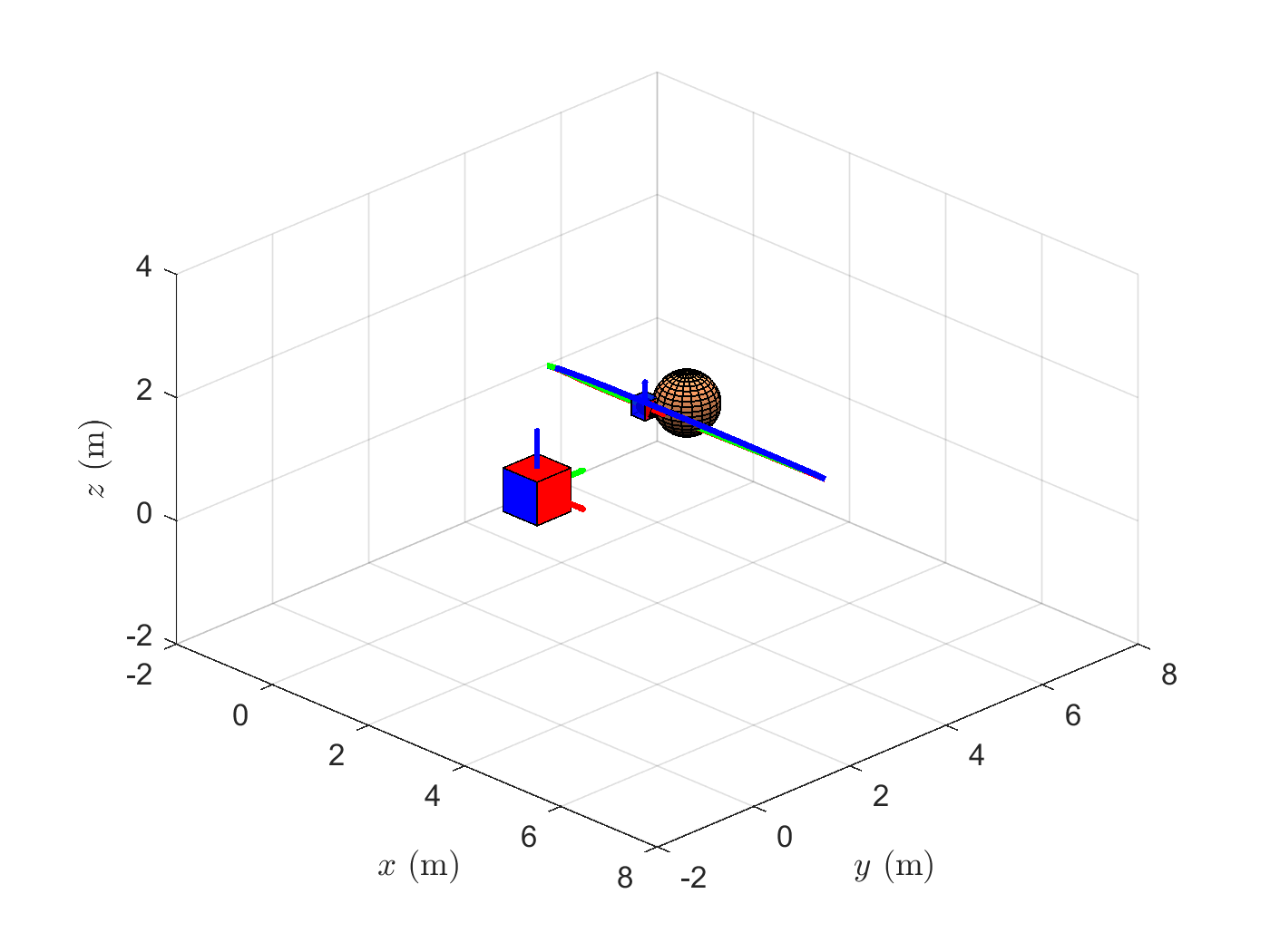}
      \label{fig:aa3}} 
\subfigure[Assembly of the final structure from the safe area]{
\includegraphics[keepaspectratio,trim={0cm .00cm 0cm .0cm},clip,width=.48\textwidth]{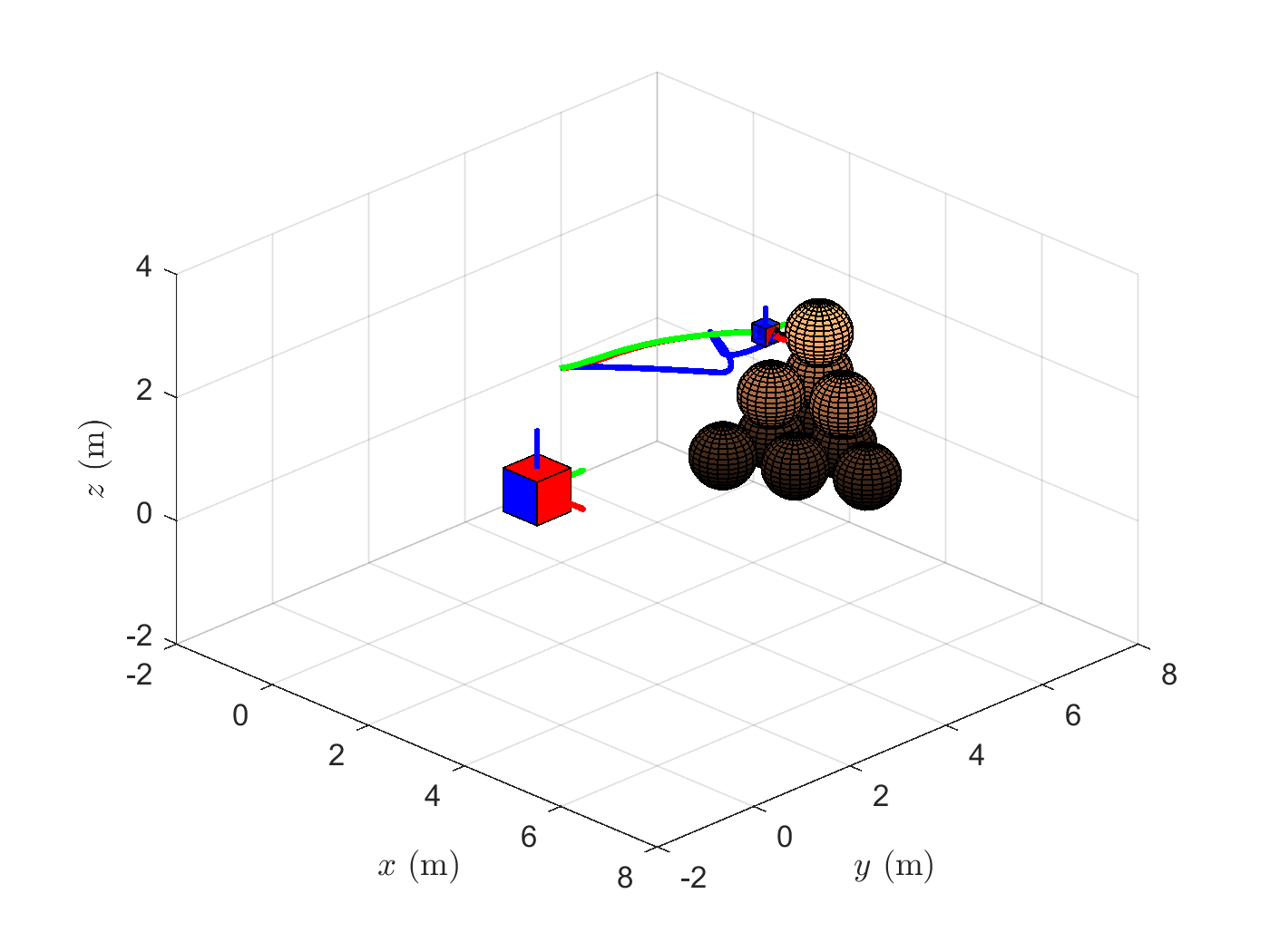}
      \label{fig:aa4}} 
        \caption{Trajectory snapshots of the on-orbit assembly process using Astrobee}\label{fig:snapshot}
\end{centering}
\end{figure}

Figure \ref{fig:mpc} shows the planner, controller, and estimation responses for the Astrobee free-flyer for each of the segments in the assembly process. In Segment 1, a standard MPC method (green) is used to track the sampling-based planner (red) for Astrobee with known inertial parameters as shown in Figure \ref{fig:mpcseg1}. The control constraints are limited to $\pm.5$ N. The snapshot of this path is given by Figure \ref{fig:aa1} where blue is the sampling-based trajectory obtained from LQR-RRT* and LQR shortcutting and green is the standard MPC response. Once the Astrobee reaches the 3D printer, it manipulates the ellipsoid 3D printed object, and the inertial estimation process occurs in Segment 2.

In Segment 2, the inertial parameters of the Astrobee with object system are unknown. Thus, inertial estimates are found by obtaining an information rich trajectory between the 3D printer and a safe area. This is given by the red line in Figure \ref{fig:mpcseg2}. Note that in our implementation, the information rich trajectory found by maximizing the FIM does not naturally consider collision avoidance, so a safe area is chosen to prevent obstacle collisions during the inertial estimation process. With the information rich plan, robust tube MPC (green) is used to track the path while considering the inertial uncertainty in the system as shown in Figure \ref{fig:mpcseg2}. The snapshot of the controlled response (green) and the information rich trajectory (red) is given in Figure \ref{fig:aa2}. Once the information-rich trajectory has been tracked, the data gathered is used for batch estimation of the inertial parameters. Control inputs provided to the robot as well as measurement data (here, pose estimates from a localization algorithm) are used for parameter estimation. Fig. \ref{fig:inertiaseg2} shows the progression of the mass estimates and their covariance with each iteration. Plans are in place to speed up the batch processing computation time by employing sparse incremental approaches \cite{kaess2008isam}.

With the inertial properties known, the Astrobee with object moves from the safe area to the assembly area for construction in Segment 3 given by Figures \ref{fig:mpcseg3} and \ref{fig:aa3}. The snapshot in Figure \ref{fig:aa3} shows the path the Astrobee with object takes from the safe area to the assembly area where the blue and green lines are the sampling-based plan and the robust tube MPC response, respectively. For Figure \ref{fig:mpcseg3}, the red line is the sampling-based planner based off the inertial estimates. Robust tube MPC with the inertial estimates is applied to control the system to the assembly area given by the green line. The responses from standard MPC and Robust tube MPC are shown without the updated inertial estimates given by the cyan and magenta lines, respectively. In these two cases, the response either overshoots the planned trajectory and saturates the control inputs quickly, or produces a more conservative response than robust tube MPC with inertial estimates. For the standard MPC without inertial estimates case (cyan), this response can lead to collisions near the assembly or 3D printer area. For the robust tube MPC without inertial estimates case (magenta), the trajectory has a slower response than the robust tube MPC with inertial estimates counterpart. This is because the covariance of the unknown inertial properties are much higher than the robust tube MPC response with inertia estimates, thus significant more control authority is applied to mitigate inertial uncertainty to promote control robustness. In either case, knowing the inertial properties lead to a more accurate system model for control. Figure \ref{fig:aa4} shows the completed assembly process of the structure using the Astrobee free-flyer where the green and blue lines are the sampling-based plan and the robust tube MPC response, respectively. From this figure, the Astrobee with object system was able to converge onto the target state for assembly without collisions with the structure by applying a safety factor in $P_{obs}$ in Eq. \eqref{ellcon}. By considering and estimating the inertial properties of the assembly system, control of collision-free trajectories for on-orbit assembly was found.


%

\section{Conclusion}
The objective of the paper is to formulate the motion planning and control of a robotic free-flyer for on-orbit assembly in the presence of inertial uncertainty due to object manipulation. By extending LQR-RRT*, LQR shortcutting, obstacle avoidance, inertial model learning, and robust tube MPC to the Astrobee testbed, safe and uncertainty-aware techniques are obtained for on-orbit robotic assembly of structures characterized by ellipsoidal keep-out zones, which can be expanded to constructing next generation telescopes, space stations, and communication satellites. Further work will entail physical experiments with the Astrobee free-flyer on the ISS and characterizing contact during the manipulation.
\section{Acknowledgment}
This research was supported by an appointment to the Intelligence Community Postdoctoral Research Fellowship Program at Massachusetts Institute of Technology, administered by Oak Ridge Institute for Science and Education through an interagency agreement between the U.S. Department of Energy and the Office of the Director of National Intelligence. Funding for this work was provided by the NASA Space Technology Mission Directorate through a NASA Space Technology Research Fellowship under grant 80NSSC17K0077. This work was also supported by the Portuguese Science Foundation (FCT) grant PD/BD/150632/2020, the LARSyS - FCT Plurianual funding 2020-2023, and an MIT Seed Project under the MIT Portugal Program. The authors gratefully acknowledge the support that enabled this research.

\bibliographystyle{AAS_publication}   
\bibliography{references}   

\begin{thebibliography}{10}

\bibitem{goward2001landsat}
S.~N. Goward, J.~G. Masek, D.~L. Williams, J.~R. Irons, and R.~Thompson, ``The
  Landsat 7 mission: Terrestrial research and applications for the 21st
  century,''  {\em Remote Sensing of Environment}, Vol.~78, No.~1-2, 2001,
  pp.~3--12.

\bibitem{reed2016restore}
B.~B. Reed, R.~C. Smith, B.~J. Naasz, J.~F. Pellegrino, and C.~E. Bacon, ``The
  restore-L servicing mission,''  {\em AIAA space 2016}, p.~5478, 2016.

\bibitem{coll2020satellite}
G.~T. Coll, G.~Webster, O.~Pankiewicz, K.~Schlee, T.~Aranyos, B.~Nufer,
  J.~Fothergill, G.~Tamasy, M.~Kandula, A.~Felt, {\em et~al.}, ``Satellite
  Servicing Projects Division Restore-L Propellant Transfer Subsystem Progress
  2020,''  {\em AIAA Propulsion and Energy 2020 Forum}, 2020, p.~3795.

\bibitem{saleh2003flexibility}
J.~H. Saleh, E.~S. Lamassoure, D.~E. Hastings, and D.~J. Newman, ``Flexibility
  and the value of on-orbit servicing: New customer-centric perspective,''
  {\em Journal of Spacecraft and Rockets}, Vol.~40, No.~2, 2003, pp.~279--291.

\bibitem{hoyt2013spiderfab}
R.~P. Hoyt, ``SpiderFab: An architecture for self-fabricating space systems,''
  {\em AIAA Space 2013 conference and exposition}, 2013, p.~5509.

\bibitem{patane2017archinaut}
S.~Patane, E.~R. Joyce, M.~P. Snyder, and P.~Shestople, ``Archinaut: In-Space
  Manufacturing and Assembly for Next-Generation Space Habitats,''  {\em AIAA
  SPACE and astronautics forum and exposition}, 2017, p.~5227.

\bibitem{jewison2014definition}
C.~Jewison, D.~Sternberg, B.~McCarthy, D.~W. Miller, and A.~Saenz-Otero,
  ``Definition and testing of an architectural tradespace for on-orbit
  assemblers and servicers,''  2014.

\bibitem{bualat2015astrobee}
M.~Bualat, J.~Barlow, T.~Fong, C.~Provencher, and T.~Smith, ``Astrobee:
  Developing a free-flying robot for the international space station,''  {\em
  AIAA SPACE 2015 Conference and Exposition}, 2015, p.~4643.

\bibitem{rosen2007computer}
D.~W. Rosen, ``Computer-aided design for additive manufacturing of cellular
  structures,''  {\em Computer-Aided Design and Applications}, Vol.~4, No.~5,
  2007, pp.~585--594.

\bibitem{schaedler2016architected}
T.~A. Schaedler and W.~B. Carter, ``Architected cellular materials,''  {\em
  Annual Review of Materials Research}, Vol.~46, 2016, pp.~187--210.

\bibitem{thomas2017effect}
D.~Thomas, M.~P. Snyder, M.~Napoli, E.~R. Joyce, P.~Shestople, and T.~Letcher,
  ``Effect of Acrylonitrile Butadiene Styrene Melt Extrusion Additive
  Manufacturing on Mechanical Performance in Reduced Gravity,''  {\em AIAA
  SPACE and astronautics forum and exposition}, 2017, p.~5278.

\bibitem{petersen2011termes}
K.~H. Petersen, R.~Nagpal, and J.~K. Werfel, ``Termes: An autonomous robotic
  system for three-dimensional collective construction,''  {\em Robotics:
  science and systems VII}, 2011.

\bibitem{willmann2012aerial}
J.~Willmann, F.~Augugliaro, T.~Cadalbert, R.~D'Andrea, F.~Gramazio, and
  M.~Kohler, ``Aerial robotic construction towards a new field of architectural
  research,''  {\em International journal of architectural computing}, Vol.~10,
  No.~3, 2012, pp.~439--459.

\bibitem{jenett2019material}
B.~Jenett, A.~Abdel-Rahman, K.~Cheung, and N.~Gershenfeld, ``Material--Robot
  System for Assembly of Discrete Cellular Structures,''  {\em IEEE Robotics
  and Automation Letters}, Vol.~4, No.~4, 2019, pp.~4019--4026.

\bibitem{dogar2015multi}
M.~Dogar, R.~A. Knepper, A.~Spielberg, C.~Choi, H.~I. Christensen, and D.~Rus,
  ``Multi-scale assembly with robot teams,''  {\em The International Journal of
  Robotics Research}, Vol.~34, No.~13, 2015, pp.~1645--1659.

\bibitem{werfel2014designing}
J.~Werfel, K.~Petersen, and R.~Nagpal, ``Designing collective behavior in a
  termite-inspired robot construction team,''  {\em Science}, Vol.~343,
  No.~6172, 2014, pp.~754--758.

\bibitem{badawy2008orbit}
A.~Badawy and C.~R. McInnes, ``On-orbit assembly using superquadric potential
  fields,''  {\em Journal of Guidance, Control, and Dynamics}, Vol.~31, No.~1,
  2008, pp.~30--43.

\bibitem{perez2012lqr}
A.~Perez, R.~Platt, G.~Konidaris, L.~Kaelbling, and T.~Lozano-Perez,
  ``LQR-RRT*: Optimal sampling-based motion planning with automatically derived
  extension heuristics,''  {\em 2012 IEEE International Conference on Robotics
  and Automation}, IEEE, 2012, pp.~2537--2542.

\bibitem{geraerts2007creating}
R.~Geraerts and M.~H. Overmars, ``Creating high-quality paths for motion
  planning,''  {\em The international journal of robotics research}, Vol.~26,
  No.~8, 2007, pp.~845--863.

\bibitem{sathya2018embedded}
A.~Sathya, P.~Sopasakis, R.~Van~Parys, A.~Themelis, G.~Pipeleers, and
  P.~Patrinos, ``Embedded nonlinear model predictive control for obstacle
  avoidance using PANOC,''  {\em 2018 European control conference (ECC)}, IEEE,
  2018, pp.~1523--1528.

\bibitem{doerr2020motion}
B.~Doerr and R.~Linares, ``Motion Planning and Control for On-Orbit Assembly
  using LQR-RRT* and Nonlinear MPC,''  {\em arXiv preprint arXiv:2008.02846},
  2020.

\bibitem{lampariello2005modeling}
R.~Lampariello and G.~Hirzinger, ``Modeling and experimental design for the
  on-orbit inertial parameter identification of free-flying space robots,''
  {\em ASME 2005 International Design Engineering Technical Conferences and
  Computers and Information in Engineering Conference}, American Society of
  Mechanical Engineers Digital Collection, 2005, pp.~881--890.

\bibitem{How2001}
J.~P. How and M.~Tillerson, ``{Analysis of the impact of sensor noise on
  formation flying control},''  {\em Proceedings of the American Control
  Conference}, Vol.~5, 2001, pp.~3986--3991, 10.1109/acc.2001.946298.

\bibitem{majumdar2017funnel}
A.~Majumdar and R.~Tedrake, ``Funnel libraries for real-time robust feedback
  motion planning,''  {\em The International Journal of Robotics Research},
  Vol.~36, No.~8, 2017, pp.~947--982.

\bibitem{lopez2019dynamic}
B.~T. Lopez, J.~P. Howl, and J.-J.~E. Slotine, ``Dynamic tube MPC for nonlinear
  systems,''  {\em 2019 American Control Conference (ACC)}, IEEE, 2019,
  pp.~1655--1662.

\bibitem{Webb2014}
D.~J. Webb, K.~L. Crandall, and J.~{Van Den Berg}, ``{Online parameter
  estimation via real-time replanning of continuous Gaussian POMDPs},''  {\em
  Proceedings - IEEE International Conference on Robotics and Automation},
  2014, pp.~5998--6005, 10.1109/ICRA.2014.6907743.

\bibitem{Okamoto2018}
K.~Okamoto, M.~Goldshtein, and P.~Tsiotras, ``{Optimal Covariance Control for
  Stochastic Systems under Chance Constraints},''  {\em IEEE Control Systems
  Letters}, Vol.~2, No.~2, 2018, pp.~266--271, 10.1109/LCSYS.2018.2826038.

\bibitem{Okamoto2019}
K.~Okamoto and P.~Tsiotras, ``{Optimal Stochastic Vehicle Path Planning Using
  Covariance Steering},''  {\em IEEE Robotics and Automation Letters}, Vol.~4,
  No.~3, 2019, pp.~2276--2281, 10.1109/LRA.2019.2901546.

\bibitem{Slotine}
J.-J.~E. Slotine and W.~Li, ``{Applied Nonlinear Control},''

\bibitem{espinoza2017concurrent}
A.~T. Espinoza and D.~Roascio, ``Concurrent adaptive control and parameter
  estimation through composite adaptation using model reference adaptive
  control/Kalman Filter methods,''  {\em 2017 IEEE Conference on Control
  Technology and Applications (CCTA)}, IEEE, 2017, pp.~662--667.

\bibitem{xu1994parameterization}
Y.~Xu, H.-Y. Shum, T.~Kanade, and J.-J. Lee, ``Parameterization and adaptive
  control of space robot systems,''  {\em IEEE transactions on Aerospace and
  Electronic Systems}, Vol.~30, No.~2, 1994, pp.~435--451.

\bibitem{keim2006spacecraft}
J.~A. Keim, A.~B. Acikmese, and J.~F. Shields, ``Spacecraft inertia estimation
  via constrained least squares,''  {\em 2006 IEEE Aerospace Conference}, IEEE,
  2006, pp.~6--pp.

\bibitem{murotsu1994parameter}
Y.~Murotsu, K.~Senda, M.~Ozaki, and S.~Tsujio, ``Parameter identification of
  unknown object handled by free-flying space robot,''  {\em Journal of
  guidance, control, and dynamics}, Vol.~17, No.~3, 1994, pp.~488--494.

\bibitem{ekal2018inertial}
M.~Ekal and R.~Ventura, ``On Inertial Parameter Estimation of a Free-Flying
  Robot Grasping an Unknown Object,''  {\em 2018 5th International Conference
  on Control, Decision and Information Technologies (CoDIT)}, IEEE, 2018,
  pp.~815--821.

\bibitem{yoshida2002inertia}
K.~Yoshida and S.~Abiko, ``Inertia parameter identification for a free-flying
  space robot,''  {\em AIAA Guidance, Navigation, and Control Conference and
  Exhibit}, 2002, p.~4568.

\bibitem{ma2008orbit}
O.~Ma, H.~Dang, and K.~Pham, ``On-orbit identification of inertia properties of
  spacecraft using a robotic arm,''  {\em Journal of guidance, control, and
  dynamics}, Vol.~31, No.~6, 2008, pp.~1761--1771.

\bibitem{christidi2017}
O.-O. Christidi-Loumpasefski, K.~Nanos, and E.~Papadopoulos, ``On parameter
  estimation of space manipulator systems using the angular momentum
  conservation,''  {\em 2017 IEEE International Conference on Robotics and
  Automation (ICRA)}, IEEE, 2017, pp.~5453--5458.

\bibitem{manchester2017recursive}
Z.~R. Manchester and M.~A. Peck, ``Recursive inertia estimation with
  semidefinite programming,''  {\em AIAA Guidance, Navigation, and Control
  Conference}, 2017, p.~1902.

\bibitem{ekal2020dual}
M.~Ekal and R.~Ventura, ``A Dual Quaternion-Based Discrete Variational Approach
  for Accurate and Online Inertial Parameter Estimation in Free-Flying obots,''
   {\em 2020 IEEE International Conference on Robotics and Automation (ICRA)},
  IEEE, 2020, pp.~6021--6027.

\bibitem{lichter2004state}
M.~D. Lichter and S.~Dubowsky, ``State, shape, and parameter estimation of
  space objects from range images,''  {\em IEEE International Conference on
  Robotics and Automation, 2004. Proceedings. ICRA'04. 2004}, Vol.~3, IEEE,
  2004, pp.~2974--2979.

\bibitem{vandyke2004unscented}
M.~C. VanDyke, J.~L. Schwartz, C.~D. Hall, {\em et~al.}, ``Unscented Kalman
  filtering for spacecraft attitude state and parameter estimation,''  {\em
  Advances in the Astronautical Sciences}, Vol.~118, No.~1, 2004, pp.~217--228.

\bibitem{shin1994}
J.-H. Shin and J.-J. Lee, ``Dynamic control with adaptive identification for
  free-flying space robots in joint space,''  {\em Robotica}, Vol.~12, No.~6,
  1994, pp.~541--551.

\bibitem{rackl2014parameter}
W.~Rackl and R.~Lampariello, ``Parameter identification of free-floating robots
  with flexible appendages and fuel sloshing,''  {\em Proceedings of 2014
  International Conference on Modelling, Identification \& Control}, IEEE,
  2014, pp.~129--134.

\bibitem{nanos2019parameter}
K.~Nanos and E.~Papadopoulos, ``On Parameter Estimation of Space Manipulator
  Systems with Flexible Joints Using the Energy Balance,''  {\em 2019
  International Conference on Robotics and Automation (ICRA)}, IEEE, 2019,
  pp.~3570--3576.

\bibitem{liu2018modeling}
Y.~Liu, Y.~Fu, W.~He, and Q.~Hui, ``Modeling and observer-based vibration
  control of a flexible spacecraft with external disturbances,''  {\em IEEE
  Transactions on Industrial Electronics}, Vol.~66, No.~11, 2018,
  pp.~8648--8658.

\bibitem{taylor1991parameter}
L.~W. Taylor~Jr, ``Parameter estimation for distributed parameter models of
  complex, flexible structures,''  {\em IFAC Proceedings Volumes}, Vol.~24,
  No.~3, 1991, pp.~1155--1160.

\bibitem{markley2014fundamentals}
F.~L. Markley and J.~L. Crassidis, {\em Fundamentals of spacecraft attitude
  determination and control}, Vol.~33.
\newblock Springer, 2014.

\bibitem{van1978computing}
C.~Van~Loan, ``Computing integrals involving the matrix exponential,''  {\em
  IEEE transactions on automatic control}, Vol.~23, No.~3, 1978, pp.~395--404.

\bibitem{13}
M.~Inaba and P.~Corke, eds., {\em Robotics Research}.
\newblock Springer International Publishing, 2016.

\bibitem{paden2016survey}
B.~Paden, M.~{\v{C}}{\'a}p, S.~Z. Yong, D.~Yershov, and E.~Frazzoli, ``A survey
  of motion planning and control techniques for self-driving urban vehicles,''
  {\em IEEE Transactions on intelligent vehicles}, Vol.~1, No.~1, 2016,
  pp.~33--55.

\bibitem{kallmann2008planning}
M.~Kallmann, A.~Aubel, T.~Abaci, and D.~Thalmann, ``Planning collision-free
  reaching motions for interactive object manipulation and grasping,''  {\em
  ACM SIGGRAPH 2008 classes}, pp.~1--11, 2008.

\bibitem{31}
R.~Bellman {\em et~al.}, ``The theory of dynamic programming,''  {\em Bulletin
  of the American Mathematical Society}, Vol.~60, No.~6, 1954, pp.~503--515.

\bibitem{karaman2011sampling}
S.~Karaman and E.~Frazzoli, ``Sampling-based algorithms for optimal motion
  planning,''  {\em The international journal of robotics research}, Vol.~30,
  No.~7, 2011, pp.~846--894.

\bibitem{hauser2010fast}
K.~Hauser and V.~Ng-Thow-Hing, ``Fast smoothing of manipulator trajectories
  using optimal bounded-acceleration shortcuts,''  {\em 2010 IEEE international
  conference on robotics and automation}, IEEE, 2010, pp.~2493--2498.

\bibitem{jewison2015model}
C.~Jewison, R.~S. Erwin, and A.~Saenz-Otero, ``Model predictive control with
  ellipsoid obstacle constraints for spacecraft rendezvous,''  {\em
  IFAC-PapersOnLine}, Vol.~48, No.~9, 2015, pp.~257--262.

\bibitem{jewison2017guidance}
C.~M. Jewison, {\em Guidance and control for multi-stage rendezvous and docking
  operations in the presence of uncertainty}.
\newblock PhD thesis, Massachusetts Institute of Technology, 2017.

\bibitem{Mayne2011}
D.~Q. Mayne, E.~C. Kerrigan, and P.~Falugi, ``{Robust model predictive control:
  Advantages and disadvantages of tube-based methods},''  {\em IFAC Proceedings
  Volumes (IFAC-PapersOnline)}, Vol.~44, No.~1 PART 1, 2011, pp.~191--196,
  10.3182/20110828-6-IT-1002.01893.

\bibitem{Mayne2005}
D.~Q. Mayne, M.~M. Seron, and S.~V. Rakovi{\'{c}}, ``{Robust model predictive
  control of constrained linear systems with bounded disturbances},''  {\em
  Automatica}, Vol.~41, No.~2, 2005, pp.~219--224,
  10.1016/j.automatica.2004.08.019.

\bibitem{Kolmanovsky1998}
I.~Kolmanovsky and E.~G. Gilbert, ``{Theory and computation of disturbance
  invariant sets for discrete-time linear systems},''  1998,
  10.1155/S1024123X98000866.

\bibitem{Limon2008}
D.~Limon, I.~Alvarado, T.~Alamo, and E.~Camacho, ``{On the design of robust
  tube-based MPC for tracking},''  {\em IFAC Proceedings Volumes
  (IFAC-PapersOnline)}, Vol.~17, No.~1 PART 1, 2008, pp.~15333--15338,
  10.3182/20080706-5-KR-1001.3054.

\bibitem{Rakovic2008a}
S.~V. Rakovi{\'{c}}, ``{Tube Model Predictive Control Using Homothety {\&}
  Invariance},''  2008.

\bibitem{Swevers1997}
J.~Swevers, C.~Ganseman, and D.~B. T, ``{Optimal Robot Excitation and
  Identification.pdf},''  Vol.~13, No.~5, 1997, pp.~730--740.

\bibitem{wilson2014trajectory}
A.~D. Wilson, J.~A. Schultz, and T.~D. Murphey, ``{Trajectory synthesis for
  fisher information maximization},''  {\em IEEE Transactions on Robotics},
  Vol.~30, No.~6, 2014, pp.~1358--1370, 10.1109/TRO.2014.2345918.

\bibitem{crassidis2011}
J.~L. Crassidis and J.~L. Junkins, {\em {Optimal Estimation of Dynamic
  Systems}}.
\newblock 2011.

\bibitem{burri2018framework}
M.~Burri, M.~Bloesch, Z.~Taylor, R.~Siegwart, and J.~Nieto, ``A framework for
  maximum likelihood parameter identification applied on MAVs,''  {\em Journal
  of Field Robotics}, Vol.~35, No.~1, 2018, pp.~5--22.

\bibitem{kaess2008isam}
M.~Kaess, A.~Ranganathan, and F.~Dellaert, ``iSAM: Incremental smoothing and
  mapping,''  {\em IEEE Transactions on Robotics}, Vol.~24, No.~6, 2008,
  pp.~1365--1378.

\end{thebibliography}

\end{document}